%% file: main.tex
\documentclass[journal]{IEEEtran}
\usepackage{cite}

\ifCLASSINFOpdf
  \usepackage[pdftex]{graphicx}
\else
\fi
\usepackage{multirow}
\usepackage{array}
\usepackage{makecell}

\usepackage[cmex10]{amsmath}
\usepackage{url}

\input{math_commands.tex}

\usepackage{hyperref}
\usepackage{xcolor}
\usepackage{subfig}

\definecolor{myblue}{rgb}{0.043, 0.325, 0.580}
\definecolor{mygreen}{rgb}{0.219, 0.462, 0.113}

\usepackage{booktabs}
\usepackage{multirow}

\begin{document}
%
\title{Crop Classification under Varying Cloud Cover with Neural Ordinary Differential Equations}
%
%
%

\author{Nando~Metzger*\thanks{*Equal contribution}, 
        Mehmet~Ozgur~Turkoglu*,
        Stefano~D'Aronco,
        Jan~Dirk~Wegner,
        Konrad~Schindler,~\IEEEmembership{Senior~Member, ~IEEE}
\thanks{All authors are members of the EcoVision Lab, Photogrammetry and Remote Sensing, ETH Zurich; J.D. Wegner is also affiliated with the Institute for Computational Science, University of Zurich.}
}

%
%

\markboth{}%
{Shell \MakeLowercase{\textit{et al.}}: Bare Demo of IEEEtran.cls for Journals}
%



\maketitle

\begin{abstract}
Optical satellite sensors cannot see the Earth's surface through clouds. Despite the periodic revisit cycle, image sequences acquired by Earth observation satellites are therefore \emph{irregularly} sampled in time.
State-of-the-art methods for crop classification (and other time series analysis tasks) rely on techniques that implicitly assume regular temporal spacing between observations, such as recurrent neural networks (RNNs).
We propose to use neural ordinary differential equations (NODEs) in combination with RNNs to classify crop types in irregularly spaced image sequences. The resulting ODE-RNN models consist of two steps: an update step, where a recurrent unit assimilates new input data into the model's hidden state; and a prediction step, in which NODE propagates the hidden state until the next observation arrives.
The prediction step is based on a continuous representation of the latent dynamics, which has several advantages. At the conceptual level, it is a more natural way to describe the mechanisms that govern the phenological cycle. From a practical point of view, it makes it possible to sample the system state at arbitrary points in time, such that one can integrate observations whenever they are available, and extrapolate beyond the last observation.
%
Our experiments show that ODE-RNN indeed improves classification accuracy over common baselines such as LSTM, GRU, temporal convolutional network, and transformer. The gains are most prominent in the challenging scenario where only few observations are available (i.e., frequent cloud cover). Moreover, we show that the ability to extrapolate translates to better classification performance early in the season, which is important for forecasting.
\end{abstract}

\begin{IEEEkeywords}
deep learning, ordinary differential equations, ODE, neural ODE, RNN, crop classification, time-series
\end{IEEEkeywords}

%
\IEEEpeerreviewmaketitle

\section{Introduction}
\label{Introduction}
\input{01_intro}

\section{Related Work}
\label{Related Work}

\input{02_related_work}

\section{Method}
\label{Method}
\input{03_background}
\input{04_method}

\input{05_experiments}

\section{Conclusion}
\label{Conclusion}
\input{06_conclusion}

\section*{Acknowledgments}
This research has been funded, in part, by the Swiss Federal Office for Agriculture (FOAG), through project \emph{DeepField}.

\bibliographystyle{IEEEtran}
\bibliography{IEEEabrv,bib}

\input{biographies/bios}

\clearpage
\appendices
\input{08_appendix}

\end{document}

%% file: math_commands.tex
\usepackage{amsmath,amsfonts,bm}

\def\eqref#1{equation~\ref{#1}}

\def\1{\bm{1}}

\def\vb{{\bm{b}}}
\def\vc{{\bm{c}}}

\def\vf{{\bm{f}}}

\def\vh{{\bm{h}}}
\def\vi{{\bm{i}}}

\def\vo{{\bm{o}}}

\def\vr{{\bm{r}}}

\def\vx{{\bm{x}}}
\def\vy{{\bm{y}}}
\def\vz{{\bm{z}}}

\def\mW{{\bm{W}}}

\DeclareMathAlphabet{\mathsfit}{\encodingdefault}{\sfdefault}{m}{sl}
\SetMathAlphabet{\mathsfit}{bold}{\encodingdefault}{\sfdefault}{bx}{n}



%% file: 01_intro.tex

\IEEEPARstart{M}{onitoring} of agricultural lands is important to  manage food production, biodiversity and forestry, among others. An increasing world population and changes in consumer habits require new agricultural areas or intensification of existing ones: $\approx$38\% of the Earth's land surface are already covered with crops and pastures \cite{land_use}.  Cropland expansion and more intensive agriculture are linked to ecological problems like deforestation, biodiversity loss and soil degradation. Crop monitoring supports land management to minimise such negative impacts.

With the increasing availability of high-resolution satellite data and annotations, crop classification with machine learning methods has made great progress. Like vegetation in general, agricultural crops have class-specific spectral properties depending on soil structure (soil brightness, roughness), vegetation architecture (leaf area index, leaf angle, etc.) and leaf biochemistry (chlorophyll, water content, nitrogen content, etc.).
Importantly, the reflectance of agricultural crops varies strongly due to their pronounced phenological cycle~\cite{temporalspectral, timespacetradeoff}, so time series modeling is essential to achieve good classification.


\begin{figure*}[t]
    \centering
    \renewcommand\tabcolsep{0pt}
    \small
    \begin{tabular}{cc}
\includegraphics[width=0.9\columnwidth]{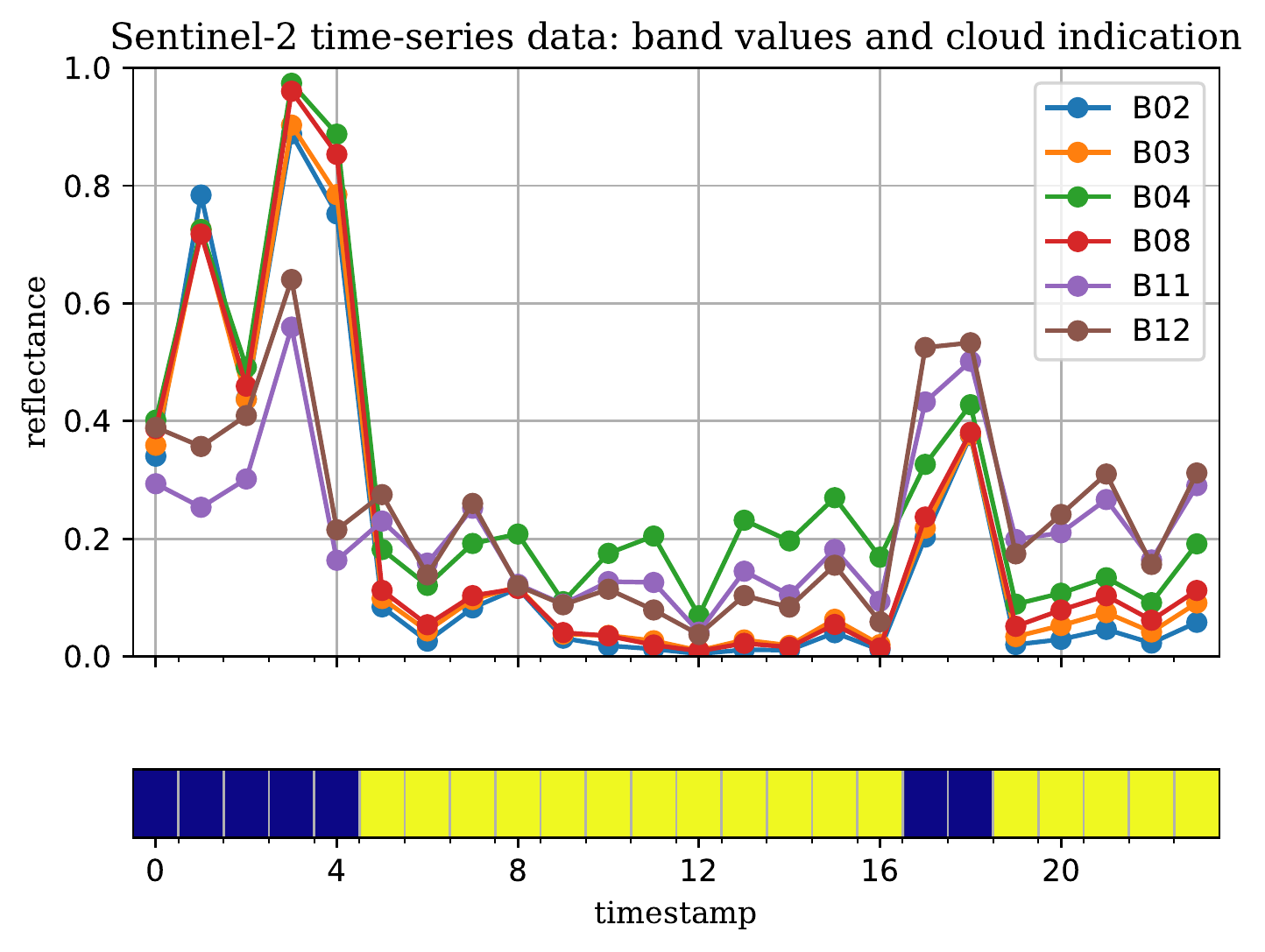} &
\includegraphics[width=0.9\columnwidth]{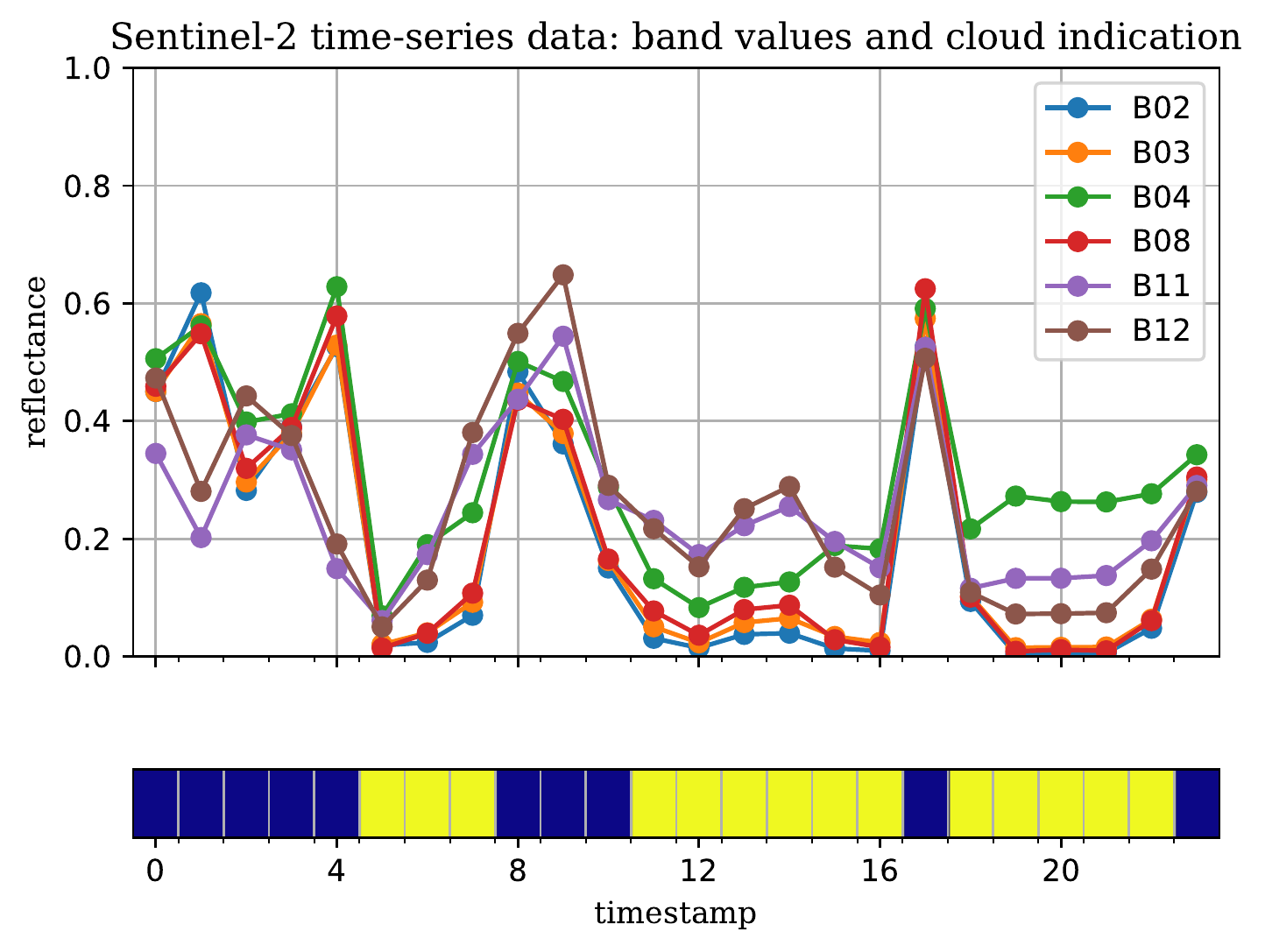}\\
\end{tabular}
\caption{Examples of Sentinel-2 time series data from TUM dataset (left for \emph{winter wheat}, right for \emph{corn}). Observations obscured by clouds are marked in blue at the bottom. Note their irregular distribution.}
\label{fig:time_series}
\end{figure*}

\begin{figure*}[t]
    \centering
        \includegraphics[width=1.7\columnwidth]{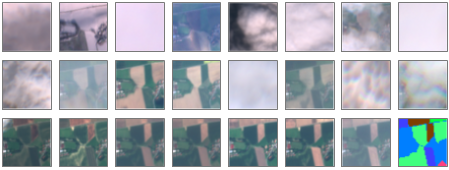}
    \caption{Example of Sentinel-2 image time series  from TUM dataset (visualisation using bands B02, B03, B04). The last image is the ground-truth map where different colors correspond to different crop types.}
    \label{fig:time_series_image}
\end{figure*}

Recurrent neural network (RNN) architectures are a powerful tool for sequence modeling, and several state-of-the-art crop mapping methods are based on RNNs \cite{marc_lstm, russwurm2018multi, fcn, star}. A main challenge when working with satellite time series in the optical spectrum (e.g, Sentinel-2 or Landsat) are atmospheric effects, in particular occlusion by clouds. If the region of interest is overcast during image acquisition, then the corresponding observations do not contain any signal from the crop on the ground, see Figs.~\ref{fig:time_series} and \ref{fig:time_series_image}.
RNNs can, in principle, learn to discard uninformative parts of the input. However, their performance degrades~\cite{gru_d,tan2020data} when uninformative observations are frequent and arbitrarily scattered across the time series as often the case in repeat-pass satellite data. For example, approximately half of all pixels in our time series dataset are invalid due to cloud cover or unavailability of a satellite image at the specified time and location.
One reason for this behaviour is that RNNs are designed for discrete data sampled at regular intervals -- in our case with equal temporal spacing. RNNs regard the individual observations as an ordered sequence, but beyond the sequence order they are unaware of time, as they do not have a "clock".
They cannot represent the dynamics of a system that is observed with variable temporal spacing, since the state update depends only on the previous state and observation, without any notion when that state is reached.

In this paper, we deal with the classification of spectral time series into crop types, in the presence of varying cloud cover.
To handle the changing and uneven data gaps due to clouds, we employ a method based on Neural Ordinary Differential Equations (NODE)~\cite{node}.
NODEs encode continuous dynamics that are, by themselves, independent of observations, and have a metric notion of time, so that observations can be included with conventional RNN operations whenever they are available.
Hence, uninformative observations can simply be skipped.
We validate the proposed method on two different datasets, in combination with two different (arguably, the two most popular) recurrent architectures, namely Long Short-Term Memory (LSTM,~\cite{lstm}) units and Gated Recurrent Units (GRU~\cite{gru}).

To the best of our knowledge, our work is the first study of neural ordinary differential equations for time  series  analysis  in  remote  sensing.
We show that the application of NODEs to satellite image time series is fairly straight-forward, opening up a host of potential applications. We believe that NODEs may be beneficial for many remote sensing tasks beyond crop mapping, since the problem of uninformative observations and/or data gaps is a ubiquitous, general problem of optical earth observation.

%% file: 02_related_work.tex
Classification of crops from satellite image data has been studied intensively in the remote sensing literature.
In early work, rule-based systems~\cite{conrad2010per} or classical machine learning~\cite{inglada2015assessment, wardlow2008large} were used. Such methods mostly relied on handcrafted features such as the Normalized Difference Vegetation Index (NDVI) \cite{foerster2012crop, ustuner2014crop, pena2011object, conrad2010per}.
To model the temporal dynamics, methods used beyond consensus voting or concatenation, also employed floating windows \cite{conrad2014derivation}, hidden Markov models \cite{siachalou2015hidden,bailly2018crop} and dynamic time warping \cite{belgiu2018sentinel}.

More recent approaches tend to avoid handcrafted features and employ deep learning.
\cite{marc_lstm} investigate the use of Long Short-Term Memory (LSTM) networks to model the temporal evolution of different crop types' spectral characteristics.
In follow-up work a convolutional version of LSTM is used to learn spatio-temporal patterns of the crop classes~\cite{russwurm2018multi}. Other popular neural architectures have also been used to represent temporal dependencies; such as temporal convolutional networks (TCNs), which employ convolutions along the time dimension \cite{pelletier2019temporal}, and transformers, which instead use a learned attention mechanism to focus on informative time steps in the sequence~\cite{breizhcrops,russwurm2020self, pixel_set,ltae}.

A recurring question when working with time series of observations is how to handle missing data.
A frequent strategy is imputation, i.e., explicitly filling the gaps by predicting the missing values. Many different variants exist, including simple (e.g., linear or spline) interpolation, supervised learning methods such as $k$-nearest neighbours \cite{batista2002study} or Random Forests \cite{stekhoven2012missforest}, and algebraic methods for low-rank matrix completion~\cite{koren2009matrix,mazumder2010spectral}.
These methods have also been used in remote sensing. In~\cite{nguyen2018comparison} the authors used a $k$-NN imputation techniques to estimate the forest biomass from Landsat time series, similarly \cite{brooks2012fitting} predicted the missing data with by leveraging Fourier analysis.
%

There have been attempts to extend recurrent networks and propagate information through data gaps. In the simplest case missing values are replaced by the last observed value or the dataset mean, but it has also been proposed to gradually decay the input towards the mean~\cite{gru_d}.
That work also suggests to append the time that has passed since the previous observation to the input.
\cite{tresp1998solution} 
integrate an RNN with a Kalman filter, whereas GRU-D uses exponential decays to model the behavior of the variables over time.
For remote sensing images it has been proposed to reconstruct the missing data with a convolutional encoder-decoder network, exploiting the fact that gaps with associated ground truth are trivial to simulate~\cite{zhang2018missing} for training.

Neural Ordinary Differential Equations (NODEs)~\cite{node} can be used as a generic building block to integrate a self-contained dynamic model into deep learning architectures, see below. By themselves, NODEs do not ingest observations, but they can be combined with standard recurrent units to form ODE-RNNs, where a NODE represents the dynamics and some RNN unit serves to update the hidden state when input is available at irregular intervals~\cite{latent_ode}. Extensions of that idea include a GRU-like version of NODE, where the sporadic input of observations is interpreted as a Bayes update, termed ODE-GRU-Bayes~\cite{odegrubayes}; and using Controlled Differential Equations to obtain smooth temporal dynamics without jumps when the hidden state is updated with new input data~\cite{kidger2020neural}.

We are only aware of one other application of NODE in remote sensing. \cite{paoletti2019neural} apply a NODE-based, \emph{continuous-depth} counterpart of the ResNet convolutional, architecture for hyperspectral image classification. The authors show that their approach can improve performance while having a lower memory footprint. However, conceptually and methodologically the approach of \cite{paoletti2019neural} is very different from our proposed model here, which proposes NODE for satellite image-sequence analysis.

%% file: 03_background.tex
\subsection{Recurrent Neural Networks}
Recurrent neural networks have established themselves as a powerful tool for modelling sequential data. They have brought significant progress in a variety of applications, in recent years notably language processing 
and speech recognition~\cite{nallapati2016abstractive, speechrnn, sak2014long}.
RNNs are feed-forward neural networks that iteratively update a hidden \emph{state vector} $\vh$ with the same computational unit (or "cell").
At a given time $t$, a new observation (input) is combined with the previous state $\vh_{t-1}$ in  a non-linear mapping that outputs the new state
$\vh_t$.
\begin{equation}
\label{eq:RNNdef}
    \vh_{t} = f (\vx_{t},\vh_{t-1},\mW)\,,
\end{equation}
with $\mW$ being the trainable parameters of the cell. The input sequence has an overall length $T$, which can vary from sample to sample. The final hidden state $\vh_T$ can be seen as a "cumulative encoding" that summarises the complete sequence, and serves as input for a decoder that maps it to the desired output, for instance a class label or a forecast of some biophysical variable. 
To fit the model to the training data, its parameters $\mW$ are adjusted to minimise a suitable loss function that measures the prediction error of either the final states or all the intermediate states of the training sequences. Usually that minimisation uses some form of stochastic gradient descent \cite{taylorbackprop, backpropagation}.

\begin{figure*}[htb]
        \centering
        \begin{tabular}{ccc}
            \includegraphics[height=0.35\columnwidth]{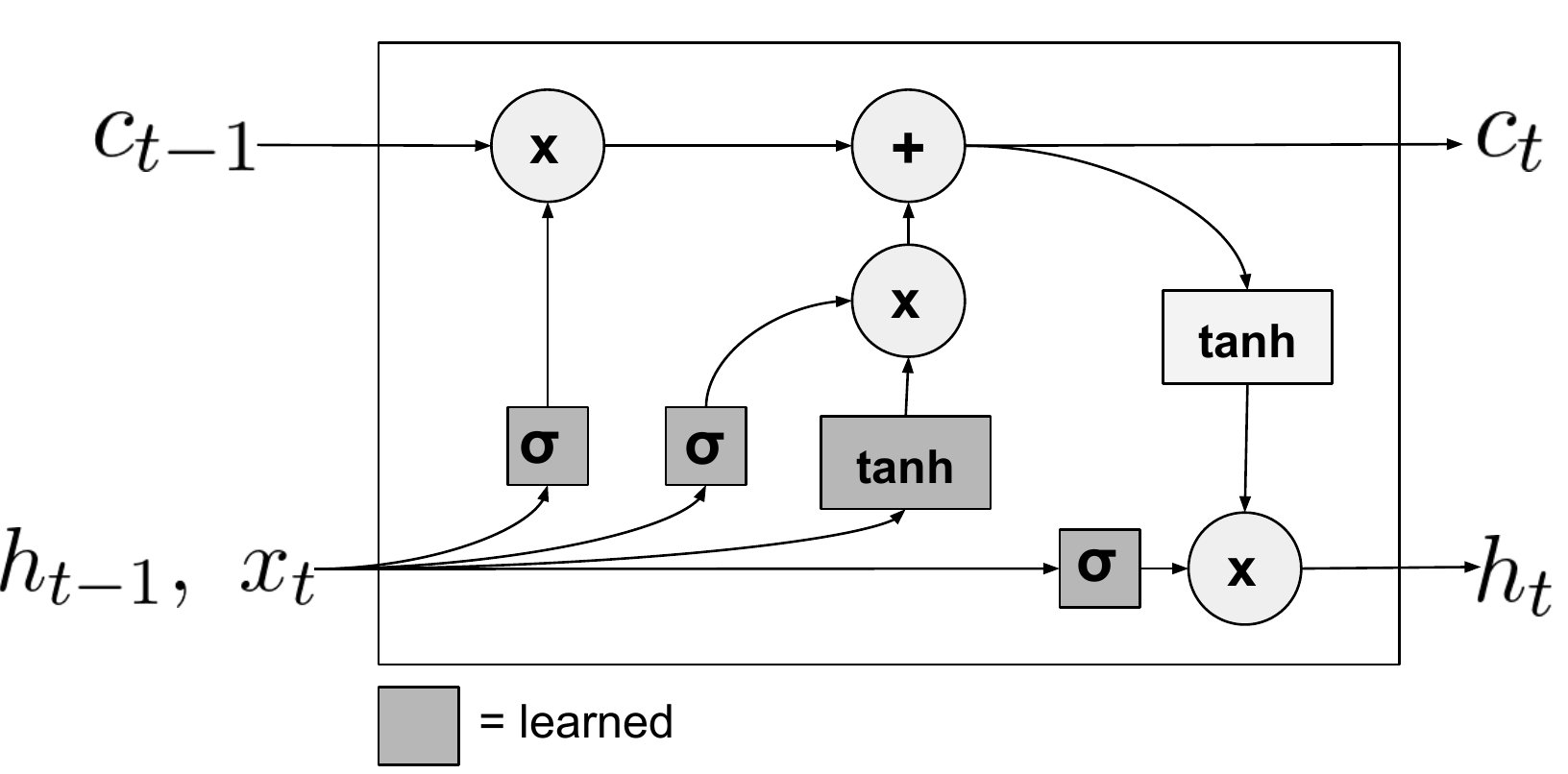} & &
            \includegraphics[height=0.35\columnwidth]{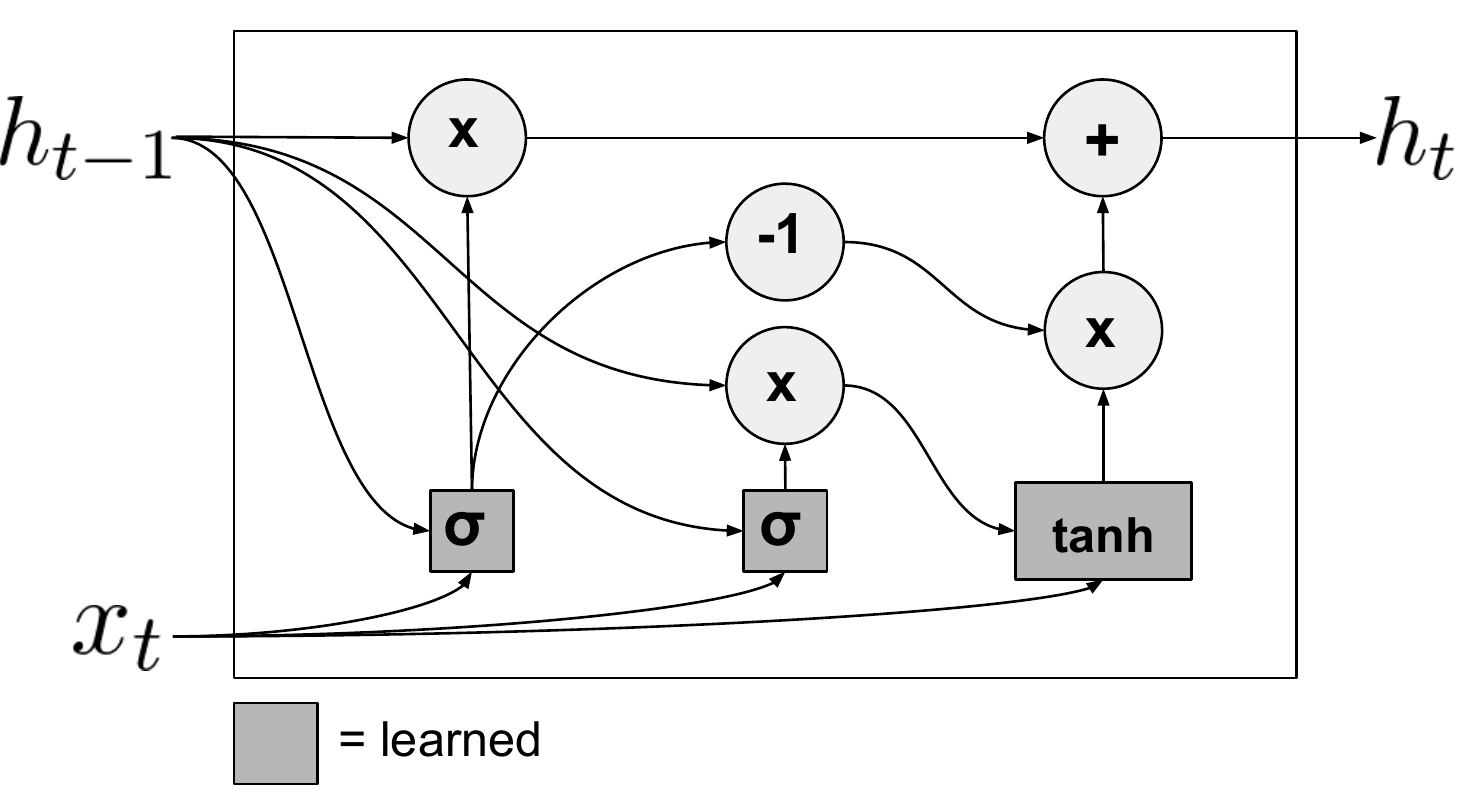} \\
\fbox{\parbox[t][2.8cm][c]{0.38\textwidth}{\small
        \begin{align*}
        &\vi_t = \sigma(\mW_{xi}\vx_t+ \mW_{hi}\vh_{t-1}+ \vb_i)\\
        &\vf_t = \sigma(\mW_{xf}\vx_t+ \mW_{hf}\vh_{t-1}+ \vb_f)\\
        &\vo_t = \sigma(\mW_{xo}\vx_t+ \mW_{ho}\vh_{t-1}+ \vb_o)\\
        &\vz_t = \tanh (\mW_{xz}\vx_t+ \mW_{hz}\vh_{t-1}+ \vb_z)\\
        &\vc_t = \vf_t \circ \vc_{t-1} + \vi_t \circ \vz_t\\ 
        &\vh_t = \vo_t \circ \tanh(\vc_t)
        \end{align*}
        }} & &
\raisebox{-0.4cm}{
\fbox{\parbox[t][2cm][c]{0.38\textwidth}{\small
        \begin{align*}
        &\vf_t = \sigma(\mW_{xf}\vx_t+ \mW_{hf}\vh_{t-1}+ \vb_f)\\
        &\vr_t = \sigma (\mW_{xr}\vx_t+ \mW_{hr}\vh_{t-1}+ \vb_r)\\ 
        &\vz_t = \tanh (\mW_{xz} \vx_{t} +\mW_{hz}(\vr_t \circ \vh_{t-1}) + \vb_z ) \\
        &\vh_t = \vf_t \circ \vh_{t-1} + (1 - \vf_t) \circ \vz_t
        \end{align*}
    }}} \\ & & \\
        (a) LSTM & & (b) GRU
\end{tabular}    
    \caption{RNN cell structures and dynamics of LSTM and GRU. 
    $\sigma$ is the Sigmoid function, $\circ$ denotes the Hadamard (element-wise) product, and $\vi_t$, $\vf_t$, $\vo_t$ are respectively the \emph{input} gate, \emph{forget} gate, and \emph{output} gate activations. $\vz_t$ is the update. LSTM additionally has a separate cell state $\vc_t$, whereas GRU has a \emph{reset} gate $\vr_t$ instead of the input and output gates.
    }
    \label{fig:dynamics}
\end{figure*}

RNNs can in principle handle sequences of arbitrary and varying length; however, in their most basic form,
\begin{equation}\vh_{t} = \tanh(\mW_{x}\vx_t+ \mW_{h}\vh_{t-1}+ \vb)
\end{equation}
 they struggle to capture long-range dependencies in the input, the error signal is diluted more and more with increasing distance (the "vanishing gradient" problem).
 To address this limitation, gated architectures have been proposed, most prominently Long Short-Term Memory (LSTM) \cite{lstm} and Gated Recurrent Units (GRU) \cite{gru}. They use \emph{gates} to store and retain information over longer time intervals, so as to mitigate the vanishing gradient problem, see Fig.~\ref{fig:dynamics}.
Gates are modulation functions that control the information flow into hidden state variables (note, LSTM has a second such variable, the "cell state"). Both LSTM and GRU are able to
capture long- and short-range dependencies in the data \cite{sutskever2014sequence, graves2013speech}.


\subsection{Neural Ordinary Differential Equations}
RNNs update their hidden state in discrete steps that are implicitly assumed to be evenly spaced, as the state change between two adjacent steps does not depend on their time difference. However, natural phenomena usually follow some underlying process that evolves continuously and longer time intervals correspond to larger changes. The hidden state of an RNN instead, cannot be sampled at arbitrary times, but only at every discrete update.
Neural Ordinary Differential Equations (NODE) as proposed by \cite{node} can be seen as a continuous version of RNNs. Instead of parameterising the update of the hidden state as in \eqref{eq:RNNdef}, its continuous dynamics is parameterised by a neural network $f_\theta$:
\begin{equation}
\label{eq:NODEdef}
    \frac{d\vh(t)}{dt} = f_{\theta} (\vh_{t}, t,\mW)
\end{equation}
\noindent
where $\theta$ are the network parameters. This differential equation corresponds to an initial value problem with input $\vh_{0}$ and solution $\vh_{T}$, hence it can be used as a feed-forward building block. The NODE approach has several desirable properties:
\begin{itemize}
    \item As it defines a continuous transformation of the latent function, it can be sampled at arbitrary, irregular times, which means that the model assumptions permit missing data points in the time series.
    \item The hidden state depends explicitly on a time parameter $t$, endowing the model with an awareness of time (respectively, rate of change), as opposed to only ordering (respectively, change per step). 
    \item One can train the model with the adjoint method~\cite{adjointtraining} instead of classical backpropagation, which is more memory efficient, since one need not store intermediate values during the forward pass. 
    \item The formulation is not tied to a specific ODE solver, so one can trade off computational cost vs.\ numerical accuracy by choosing an appropriate solver.
\end{itemize}

%% file: 04_method.tex
\subsection{Combining NODE with RNN}
\begin{figure*}[t]
    \centering
        \includegraphics[width=1.85\columnwidth]{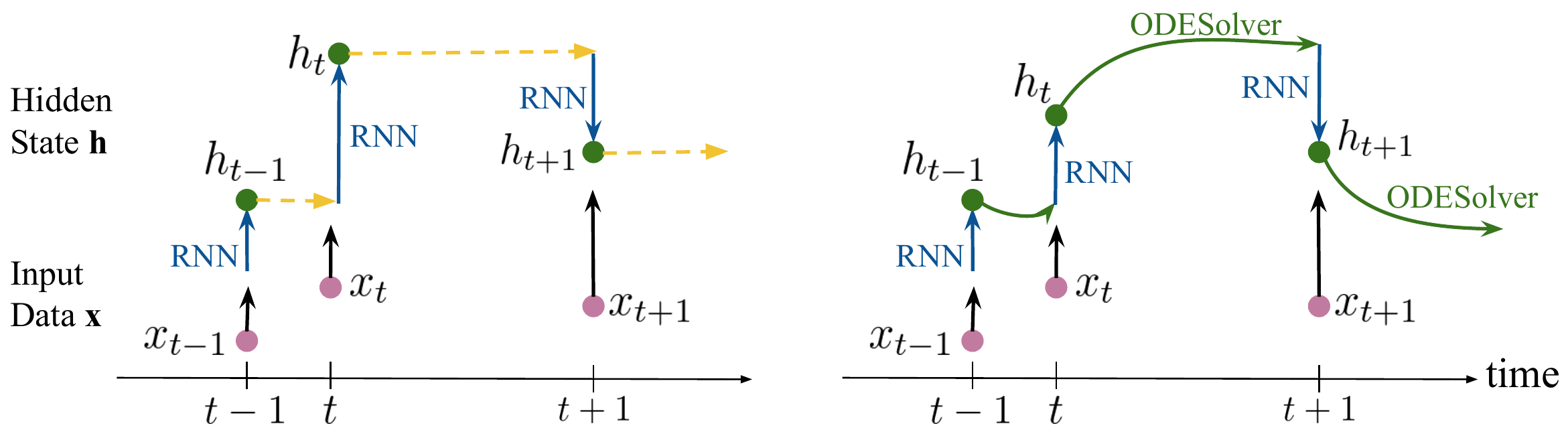}
    \caption{Illustration of the hidden state dynamics for traditional RNNs (left) and for the ODE-RNN model (right). Observations are available at time $t-1$, $t$, and $t+1$. (Left): Blue arrow shows the RNN update and the dashed yellow arrow depicts the non-continuous dynamics of RNN in the time direction. (Right): Blue arrow shows update step with RNN (\eqref{eq:update}) and green arrow shows the prediction step (time-continuous dynamics) with the ODESolver (\eqref{eq:prediction}).}
    \label{fig:model}
\end{figure*}

By itself, NODE updates the state vector $\vh$ without taking into account any observations $\vx$, i.e., it only represents the system dynamics. See \eqref{eq:NODEdef}.
To inject observations whenever they become available, we incorporate conventional recurrent units (we test both LSTM or GRU), similar to \cite{latent_ode}.
The proposed approach alternates between two steps:
\begin{enumerate}
    \item state prediction with NODE, and
    \item state update with RNN, based on the observed data.
\end{enumerate}
At time $t=0$ the hidden state $\vh$ is initialized with Gaussian noise of small magnitude, $\vh_{0} \sim \mathcal{N}(0,\,\sigma^2)\,$ with $\sigma=10^{-4}$. Starting from there, the hidden state at time $t_i$  is predicted with the ODESolver and then updated with an RNN only if an observation $\vx_{t_i}$ is available for the time step $t_i$. If no observation is available, that second step is skipped. More formally, these two steps are expressed as follows:
\begin{align}
  \text{\textcolor{mygreen}{Prediction}:}&\quad
        \vh_{t_{i}} := \mathsf{ODESolver}\big(f_{\theta},\vh_{t_{i-1}},(t_{i-1},t_{i})\big)
        \label{eq:prediction}\\
  \text{\textcolor{myblue}{Update}:}&\quad
    \vh_{t_{i}} := \mathsf{RNN}(\vh_{t_{i}},\vx_{t_{i}})
    \label{eq:update}
    \end{align}
We use a multi-layer perceptron (MLP) with two layers for the ODE network $f_\theta$, which is solved with the Euler method. The Euler method performs numerical integration with an explicit first-order approximation and is considered the simplest explicit  solver for initial value problems on ODEs.
ODEsolvers can take an infinite number of steps between two observations, depending on the step size.
In our implementation we set the number of steps equal to the timestamp difference between the current valid observation and the next valid one.
Considering the statistics, missing observations rate, of the two dataset we perform: $2.44$ steps per update for the TUM dataset (missing rate of 0.59), and $1.96$ for ZueriCrop (missing rate of 0.49).
In our experiments, we observed that a smaller step size can slightly improve the overall performance. However, improvements are not significant enough to justify the increase in computational cost.
%
%
The recurrent unit $\mathsf{RNN}$ in our experiments is a LSTM or GRU.
Figure~\ref{fig:model} schematically illustrates the evolution of the hidden state. The state $\vh_T$ at the final time $T$, after seeing all observations, serves as a feature encoding of the complete time series and is fed into a fully-connected network to map it to class probabilities for the different crop types.
%
%
%
\subsection{Learning}
The entire model, including the ODE part, is fully differentiable and trained by minimizing the following objective function:
\begin{equation}
    L = \frac{1}{N}\sum_{n=1}^{N} CE(\vy^n,\hat{\vy}^n) 
\end{equation}
where $CE$ denotes the cross-entropy loss between $\vy$ and $\hat{\vy}$ which are ground-truth and predicted crop class probabilities respectively, and $N$ is the number of time-series in the training set.
The gradient w.r.t. each model parameter $\partial L/ \partial \theta$ can be computed by backpropagation. However, a naive implementation of the chain rule is memory inefficient since all the intermediate computation of ODESolver need to be stored as well. \cite{node} shows that the gradients can be computed with low memory cost using the adjoint sensitivity method \cite{pontryagin1962mathematical} by solving a second, augmented ODE backward in time.
We use the Pytorch library \emph{torchdiffeq} \cite{torchdiffeq}, which enables memory-efficient backpropagation through the ODESolver.
\subsection{Regularization of ODE-RNN}
\label{sec:reg}
Regularization is essential for neural networks to generalize well to the test data. In the development stage, we found that artificially omitting some observations in the input sequences during training provides a good regularization effect for the proposed method in terms of data augmentation. 
Intuitively, randomly omitting a small fraction of the observations in the input time series increases the variability in the training set in terms of temporal gaps. The advantage of this additional temporal sampling diversity outweighs the slightly sparser sampling per sequence and ultimately improves performance.
To perform this regularization, we randomly sub-sample the input sequence along the time direction with factor $p\!\in\![0\hdots 1]$ ($p=1$ means no regularization). At test time, the entire sequence is presented to the model, without sub-sampling. We set $p=0.75$ for both datasets. In experiments, we apply the same regularization technique for the baselines, too.
\subsection{Implementation Details}
We implement the ODE network ($f_{\theta}$ in Eq. \ref{eq:NODEdef} and \ref{eq:prediction}) as a multi-layer perceptron with two layers. The number of neurons per layer is set to $255$ for TUM, respectively to $220$ for ZueriCrop, reducing the capacity to account for the smaller number of classes. 
The size of the hidden state vector $\vh$ is set to $80$ for both datasets.
%
%
A single fully-connected layer (perceptron) serves as classifier on top of ODE-RNN to map the final state $\vh_T$ to a class probability.
Batch normalization \cite{batch_norm} of the hidden state is performed before that classification layer, and before every LSTM/GRU update.
The model is trained with the Adamax optimizer~\cite{adam} with mini-batch size of $500$.
%
The learning rate is initially set to $0.07$ and then decreased by a constant factor $0.9995$ after each batch iteration, corresponding to a half-life of $1400$ batches.
On the TUM dataset the model was trained for $12$ epochs or a total of $\approx6'000$ batch iterations.
For the much larger ZueriCrop dataset a single epoch of $\approx 22'000$ batches was sufficient.
All models were implemented in Pytorch. Source code is available at \scalebox{0.9}[1.0]{\texttt{https://github.com/nandometzger/ODEcrop}}. 

%% file: 05_experiments.tex
\section{Datasets}
\label{Datasets}
\subsection{TU Munich Crop Data}
The TUM dataset was generated by \cite{marc_lstm}. It comprises $\approx 400k$ time series of up to $26$ multi-spectral Sentinel-2A satellite images with a ground resolution of $10\,$m. On average, 59\% of the pixels per time series are invalid, either due to cloud cover or because no satellite image is available at the specified time and location. The data is collected over a $102\times42\,$km$^\text{2}$ area north of Munich, Germany (Fig.~\ref{fig:location}), between December 2015 and August 2016, and comes with ground truth annotations with 19 classes (18 different crop types, plus a rejection class "other").
The data has been atmospherically corrected with the standard settings of the \emph{Sen2cor} toolbox \cite{louis2016sentinel}
and comprises six bands, namely B2 (blue), B3 (green) B4 (red) B8 (near infrared) B11 and B12 (both short-wave infrared).
Lower-resolution bands are upsampled to $10\,$m with nearest-neighbour interpolation.
The intensities in each band were normalised to have mean zero and unit energy.
As input to the network we use patches of $3\times 3$ pixels centred at the pixel of interest, flattened into $54$-dimensional vectors. 
The train/test split of this dataset is $\approx 80\%/20\%$. We use $20\%$ of the training set as a validation set for hyper-parameter tuning.
\subsection{ZueriCrop Data}
\begin{figure} [t]
    \centering
    \includegraphics[width=0.8\columnwidth]{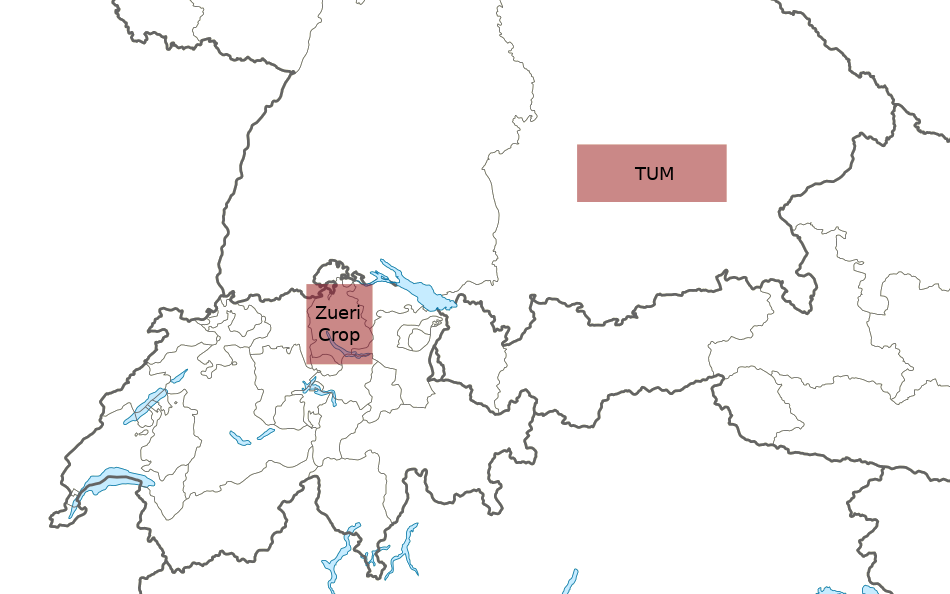}
    \caption{Location of the two datasets.
    }
    \label{fig:location}
\end{figure}

The ZueriCrop dataset was generated by \cite{turkoglu2021crop}. This dataset is collected a time series of $71$ multi-spectral Sentinel-2 satellite images with a ground resolution of $10\,$m collected over a $50\times48$ km$^\text{2}$ area over the Swiss cantons of Zurich and Thurgau (Fig.~\ref{fig:location}) between January 2019 and December 2019. On average, 49\% of the pixels per time series are invalid due to either cloud coverage or lack of satellite data at the specified time and location.
The TUM data were atmospherically corrected with the \emph{Sen2cor} toolbox \cite{louis2016sentinel} and normalised to mean zero and unit energy per channel.
We use $9$ bands, B2 (blue), B3 (green), B4 (red), B5, B6, B7 (all vegetation red edge), B8 (near infrared), B11, B12 (both short-wave infrared).
As above, lower-resolution bands are upsampled to $10\,$m with nearest-neighbour interpolation, then intensities from a $3\times3$ neighbourhood around each pixel are flattened into $81$-dimensional input vectors.
%
The original nomenclature includes a large number of crop classes with an extremely imbalanced, long-tailed distribution.
We collect all rare classes with a total of $<50'000$ pixels per class into a common rejection class "no label", which leaves us with $13$ classes for the dominant crop types.
In total, there are $3.5\cdot 10^5$ time series, split into $80\%$ for training and $20\%$ for testing. Within the training set, $20\%$ are set aside as validation set to tune the hyper-parameters.
\section{Experiments}
\label{Experiments}
\subsection{Setup}
We test two different variants of the proposed method, using either LSTM or GRU to include the observation data.
These methods are then compared to the respective baselines without the NODE part.
For quantitative performance evaluation we use the F1-score (harmonic mean between precision and recall) and the overall accuracy (fraction of correctly classified total pixels).
As a main metric we focus on the F1-score, since it is less influenced by varying class frequencies, whereas the overall accuracy tends to be biased towards the most dominant classes. We report both mean and standard deviation which are derived over three runs with different random seeds.
Uninformative observations (clouds, water) are detected with the corresponding tools of \emph{Sen2cor} \cite{louis2016sentinel} and removed from the input time series.
%
%
%

\subsection{Baselines}
\label{sec:baseline}
We compare our model to several baselines that do not employ NODE. 
The implementations of the RNN baselines (LSTM and GRU) without ODE backbone closely follow those of ODE-RNN, so as to rule out implementation differences and make them comparable.
We run three different versions of the RNN baselines: \emph{B-I} leaves out the ODE prediction step (\eqref{eq:prediction}), such that we are left with a conventional LSTM, respectively GRU model; \emph{B-II} also leaves out the ODE prediction, but adds the time interval $\delta t$ since the previous observation to the input vector. This technique gives the models the chance to learn how to handle varying delays between observations. Empirically this variant, denoted by the suffix $-\delta t$, does slightly improve the performance for both LSTM and GRU. 
%
%
Another technique to introduce temporal information into models is to add a position encoding (PE) of periodic functions, which recently became popular with the transformer architecture \cite{transformer}. We implement baseline \emph{B-III} by adding the following position encoding to the input:
\begin{align}
PE(t,i) = \sin\big(\text{day}(t)/\tau^{\frac{2i}{d}} + \frac{\pi}{2}\text{mod}(i,2)\big)
\end{align}
where $d$ is the dimension of the input ($d=54$ and $d=81$ for TUM and ZueriCrop datasets, respectively.), and $i$ the coordinates of the positional encoding. By following \cite{pixel_set} we set $\tau=1000$. $\text{day}(t)$ represents the number of days since the first observation for the observation at time $t$. In the experiments, we observe that this variant is more effective than \emph{B-II}.
A separate hyper-parameter search was done for each model. We found that, for both \emph{B-I}, \emph{B-II}, and \emph{B-III} for both GRU and LSTM, a hidden state of size $150$ worked best. 
The detailed list of hyper-parameters are given in the Appendix.
As optimiser for the baseline models, we used Adam~\cite{adam} with an initial learning rate of 0.07 and batch size of 300, which we empirically found to work best. The learning rate is decreased by a constant factor of 0.9995 after each batch iteration, corresponding to a half-life of 1400 batches.

We also compare the proposed approach with non-recurrent methods that have proven to be effective for crop mapping: (i) \cite{pelletier2019temporal} propose a temporal convolutional network (TCN) approach and (ii) \cite{russwurm2020self, pixel_set,ltae} propose a Transformer-based approach. We run \cite{ltae}'s model since they achieve state-of-the-art performance for an object-based crop mapping task on the Sentinel2-Agri dataset~\cite{pixel_set}. We use the same set of hyper-parameters as suggested in the corresponding papers.

%


\subsection{Performance Comparison}
\begin{figure*}[t]
    \centering
    \subfloat
    {{\includegraphics[height=0.23\textheight]{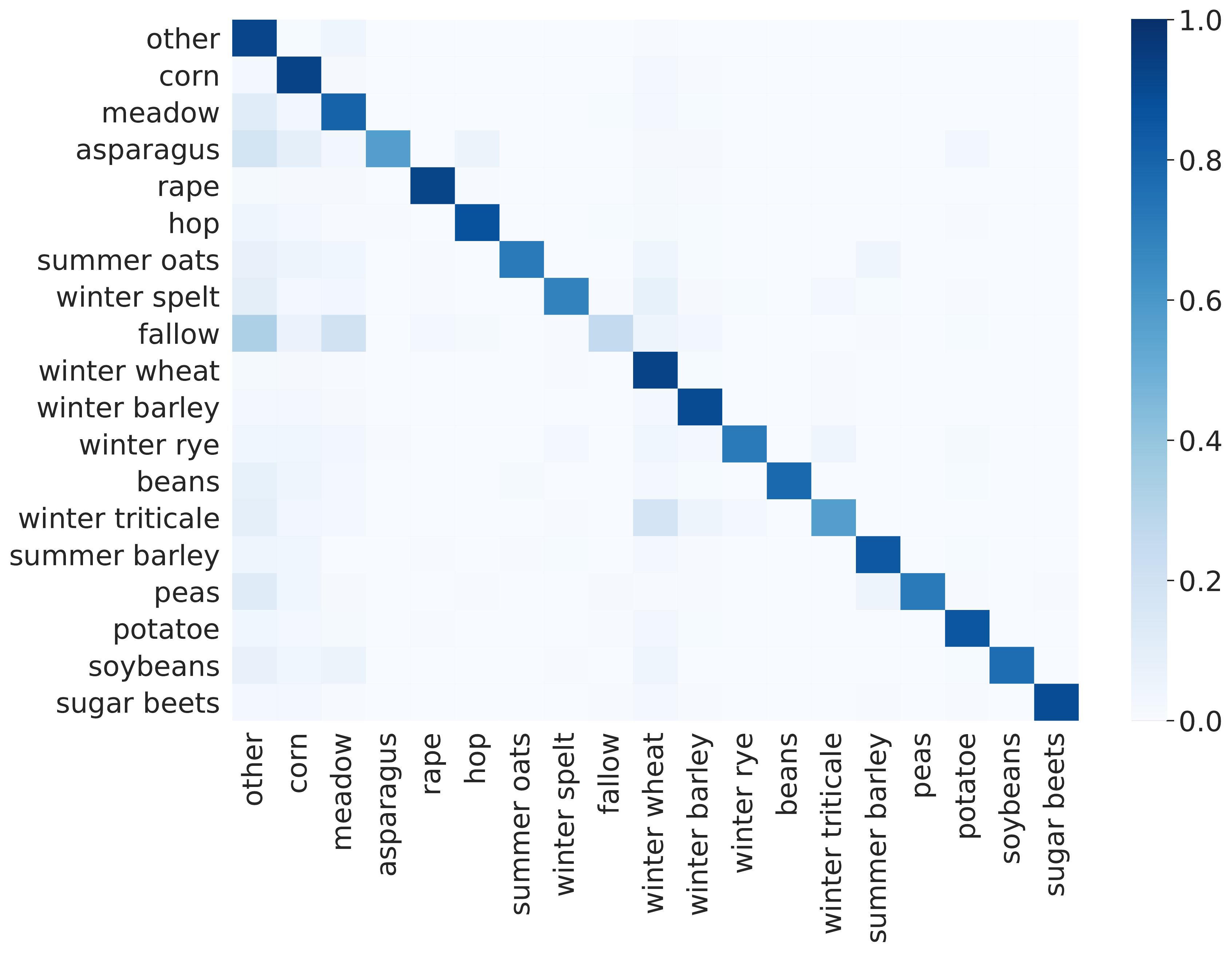} }}%
    \qquad
    \subfloat
    {{\includegraphics[height=0.23\textheight]{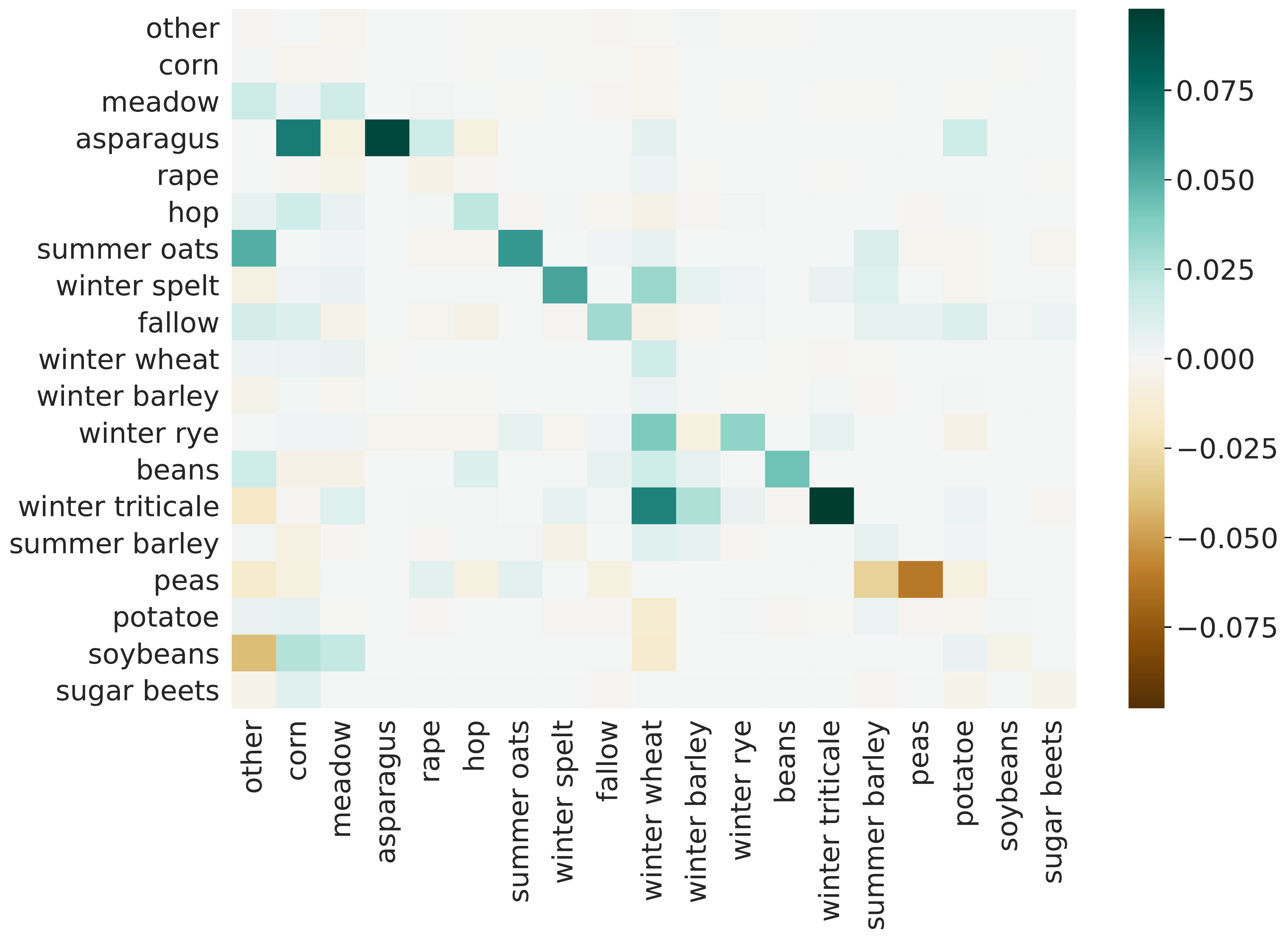} }}%
    \caption{Per-class results for TUM data. \textbf{Left: }Normalised confusion matrix of ODE-GRU. Rows denote the true classes, columns the predicted ones. \textbf{Right: }Benefit of ODE-GRU over GRU-$\delta t$. Green denotes margins in favour of ODE-GRU (higher correctness on the diagonal, respectively lower confusion off the diagonal), brown denotes margins in favour of GRU-$\delta t$.  }
    \label{fig:cm}
\end{figure*}

Quantitative classification results for the TUM and ZueriCrop datasets are reported in Tables \ref{table:result_tum} and \ref{table:result_swissdata}, respectively.
The proposed NODE method brings a consistent improvement over pure RNN models, across both performance metrics. 
Specifically, ODE-GRU outperforms all the other methods including TCN~\cite{pelletier2019temporal} and Transformer~\cite{ltae} on both datasets.
Our NODE method reduces the model size (i.e., number of parameters), and memory requirements on the GPU for training and even slightly reduces the CPU run-time for the forward pass (Tab.~\ref{table:params}).
%
%
%
%
%
For a more detailed picture, we show the full confusion matrix of ODE-GRU for the TUM data in Fig.~\ref{fig:cm}. Note the consistently high accuracy across almost all crop classes. Significant miss-classifications happen only between the most ambiguous classes, namely: fallows (i.e., unproductive areas \emph{not} currently used for crops) are confused with meadows and with the diffuse "other" class; and different types of winter grains with similar phenotype and similar phenological cycle are sometimes confused.
We also visualise the difference between the confusion matrix of ODE-GRU and that of one of the best baseline GRU-$\delta t$. For most classes, ODE-GRU increases the rate of correct classifications and reduces confusions, in some cases up to \textgreater7 percent points.
There is only a single class out of 19 where ODE-GRU performs noticeably worse (peas).

\begin{figure*}
    \centering
    \includegraphics[width=0.83\textwidth]{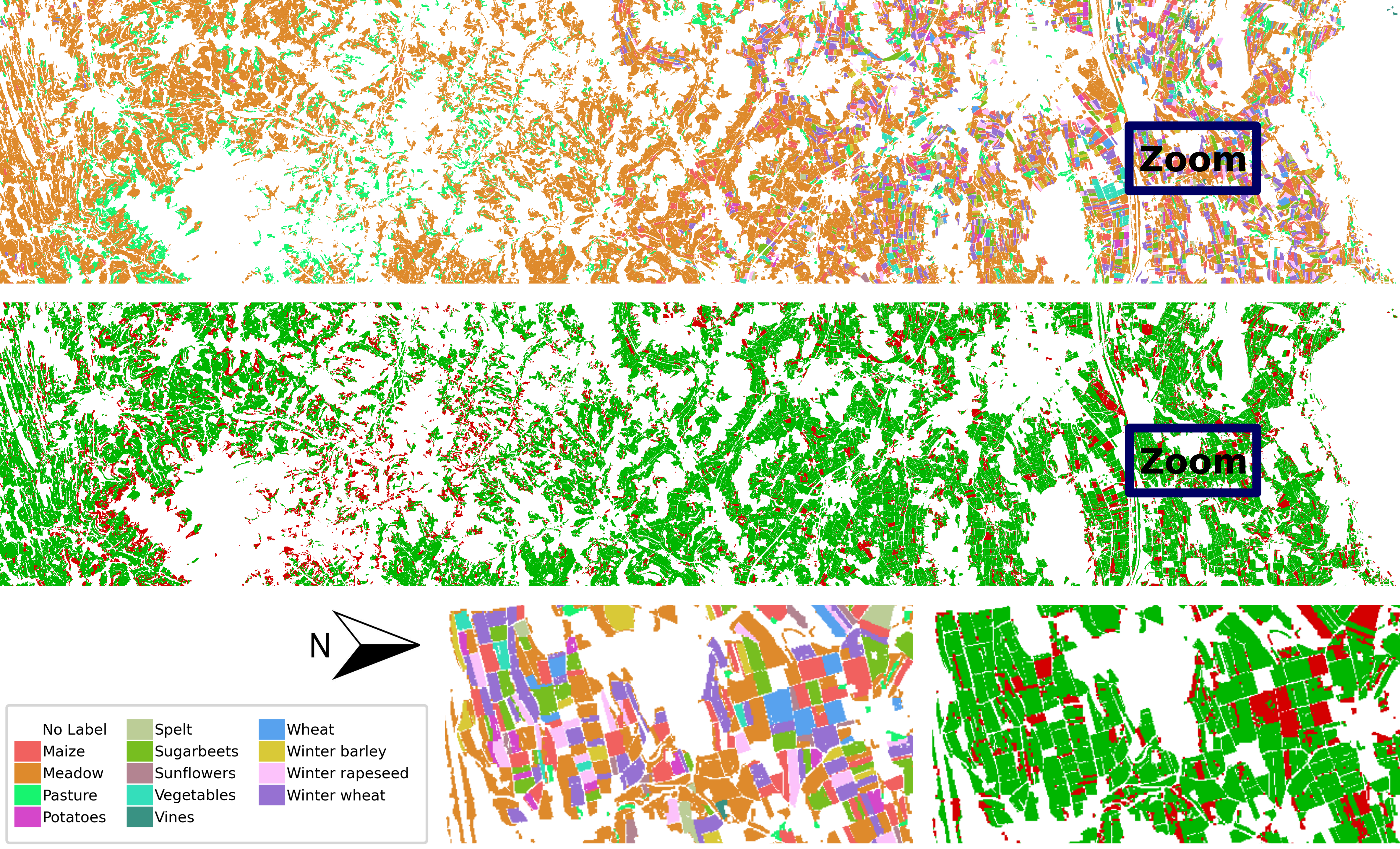}
    \caption{Qualitative result: Predicted crop map of the ODE-GRU model for the complete test area of ZueriCrop. The first map shows ground-truth pixel values where different colours correspond to different crop types. The second map illustrates the performance of the model, with correctly classified pixels in green and misclassified ones in red.}
    \label{fig:qualitative}
\end{figure*}

In Fig.~\ref{fig:qualitative}, we show qualitative results of the ODE-GRU model for the complete test area of the ZueriCrop dataset.

\begin{table}[t]
\begin{center}
\begin{tabular}{clll}
&\multicolumn{1}{c}{\bf Method} &\multicolumn{1}{c}{\bf F1-score (\%)}   &\multicolumn{1}{c}{\bf Accuracy (\%)} \\
\toprule
&\multicolumn{1}{c}{ \begin{tabular}[c]{@{}c@{}} TCN \cite{pelletier2019temporal} \end{tabular}  }  &\multicolumn{1}{c}{70.9 $\pm 0.2$} &\multicolumn{1}{c}{85.9 $\pm 0.1$}  \\
&\multicolumn{1}{c}{ \begin{tabular}[c]{@{}c@{}} Transformer \cite{ltae} \end{tabular}  }  &\multicolumn{1}{c}{79.4 $\pm 0.1$} &\multicolumn{1}{c}{88.3 $\pm 0.1$}   \\
&\multicolumn{1}{c}{ \begin{tabular}[c]{@{}c@{}} Transformer (reg.) \cite{ltae} \end{tabular}  }  &\multicolumn{1}{c}{79.8 $\pm 0.2$} &\multicolumn{1}{c}{88.3 $\pm 0.0$}   \\
\midrule
\multirow{2}{*}{
\scalebox{0.85}[1.0]{\textit{B-I}}
}
&\multicolumn{1}{c}{ \begin{tabular}[c]{@{}c@{}} LSTM \end{tabular}  }  &\multicolumn{1}{c}{76.9 $\pm 0.0$} &\multicolumn{1}{c}{87.3 $\pm 0.1$}  \\
&\multicolumn{1}{c}{ \begin{tabular}[c]{@{}c@{}} GRU \end{tabular}  }  &\multicolumn{1}{c}{78.3  $\pm 0.1$} &\multicolumn{1}{c}{87.3 $\pm 0.0$}  \\
\midrule
\multirow{3}{*}{
\scalebox{0.85}[1.0]{\textit{B-II}}
}
&\multicolumn{1}{c}{ \begin{tabular}[c]{@{}c@{}} LSTM-$\delta t$ \end{tabular}} &\multicolumn{1}{c}{77.4 $\pm 0.3$} &\multicolumn{1}{c}{87.3 $\pm 0.0$}  \\
&\multicolumn{1}{c}{ \begin{tabular}[c]{@{}c@{}} GRU-$\delta t$ \end{tabular}} &\multicolumn{1}{c}{78.2 $\pm 0.7$} &\multicolumn{1}{c}{87.6 $\pm 0.1$}   \\
&\multicolumn{1}{c}{ \begin{tabular}[c]{@{}c@{}} GRU-$\delta t$ (reg.) \end{tabular}} &\multicolumn{1}{c}{79.0 $\pm 0.7$} &\multicolumn{1}{c}{88.1 $\pm 0.1$}   \\
\midrule
\multirow{3}{*}{\textit{B-III}}
%
%
&\multicolumn{1}{c}{ \begin{tabular}[c]{@{}c@{}} LSTM-PE \end{tabular}} &\multicolumn{1}{c}{ 77.7 $\pm 0.2$} &\multicolumn{1}{c}{87.5 $\pm 0.1$}  \\
&\multicolumn{1}{c}{ \begin{tabular}[c]{@{}c@{}} GRU-PE \end{tabular}} &\multicolumn{1}{c}{ 78.6 $\pm 0.4$} &\multicolumn{1}{c}{87.8 $\pm 0.1$}   \\
&\multicolumn{1}{c}{ \begin{tabular}[c]{@{}c@{}} GRU-PE (reg.) \end{tabular}} &\multicolumn{1}{c}{79.7 $\pm 0.2$} &\multicolumn{1}{c}{88.1 $\pm 0.1$}   \\
\midrule
\multirow{3}{*}{
\scalebox{0.85}[1.0]{\textit{NODE}}
}
&\multicolumn{1}{c}{ \begin{tabular}[c]{@{}c@{}} $\quad$ODE-LSTM$\quad$ \end{tabular}}  &\multicolumn{1}{c}{78.7 $\pm 0.4$} &\multicolumn{1}{c}{87.9 $\pm 0.1$} \\
&\multicolumn{1}{c}{ \begin{tabular}[c]{@{}c@{}} ODE-GRU \end{tabular}} &\multicolumn{1}{c}{79.9 $\pm 0.2$} &\multicolumn{1}{c}{88.1 $\pm 0.0$} 
\\
&\multicolumn{1}{c}{ \begin{tabular}[c]{@{}c@{}} ODE-GRU (reg.) \end{tabular}} &\multicolumn{1}{c}{\textbf{80.4} $\pm 0.2$} &\multicolumn{1}{c}{ \textbf{88.5} $\pm 0.1$}  
\\ \bottomrule 

\end{tabular}
\end{center}
\caption{Performance comparison on TUM dataset. reg.\ stands for regularization, see Section \ref{sec:reg} and PE is positional encoding, see Section \ref{sec:baseline}. 
The best score for each metric is printed \textbf{bold}.} 
\label{table:result_tum}
\end{table}

\begin{table}[th]
\begin{center}
\begin{tabular}{clll}
&\multicolumn{1}{c}{\bf Method}  &\multicolumn{1}{c}{\bf F1-score (\%)} &\multicolumn{1}{c}{\bf Accuracy (\%)}  \\
\toprule
&\multicolumn{1}{c}{ \begin{tabular}[c]{@{}c@{}} TCN \cite{pelletier2019temporal} \end{tabular}  }     &\multicolumn{1}{c}{60.1 $\pm 0.6$} &\multicolumn{1}{c}{83.0 $\pm 0.2$}  \\
&\multicolumn{1}{c}{ \begin{tabular}[c]{@{}c@{}} Transformer \cite{ltae} \end{tabular}  }  &\multicolumn{1}{c}{71.7 $\pm 0.8$} &\multicolumn{1}{c}{85.1 $\pm 0.3$}   \\
&\multicolumn{1}{c}{ \begin{tabular}[c]{@{}c@{}} Transformer (reg.) \cite{ltae} \end{tabular}  }  &\multicolumn{1}{c}{72.6 $\pm 0.7$} &\multicolumn{1}{c}{85.4 $\pm 0.2$}  \\
\midrule
\multirow{2}{*}{
\scalebox{0.85}[1.0]{\textit{B-I}}
}
&\multicolumn{1}{c}{ \begin{tabular}[c]{@{}c@{}} LSTM \end{tabular}  }     &\multicolumn{1}{c}{ 72.2 $\pm 1.0$}  &\multicolumn{1}{c}{ 85.1 $\pm 0.2$ } \\
&\multicolumn{1}{c}{ \begin{tabular}[c]{@{}c@{}} GRU \end{tabular}  }     &\multicolumn{1}{c}{73.7  $\pm 1.0$}  &\multicolumn{1}{c}{85.3 $\pm 0.2$} \\
\midrule
\multirow{3}{*}{
\scalebox{0.85}[1.0]{\textit{B-II}}
}
&\multicolumn{1}{c}{ \begin{tabular}[c]{@{}c@{}} LSTM-$\delta  t$ \end{tabular}  }  &\multicolumn{1}{c}{ 72.7 $\pm 0.6$}    &\multicolumn{1}{c}{85.0 $\pm 0.3$}  \\
&\multicolumn{1}{c}{ \begin{tabular}[c]{@{}c@{}} GRU-$\delta  t$ \end{tabular}  }  &\multicolumn{1}{c}{74.6 $\pm 0.7$}   &\multicolumn{1}{c}{85.3 $\pm 0.1$}  \\
&\multicolumn{1}{c}{ \begin{tabular}[c]{@{}c@{}} GRU-$\delta  t$ (reg.) \end{tabular}  }  &\multicolumn{1}{c}{ 74.8 $\pm 0.4$}   &\multicolumn{1}{c}{ 85.8 $\pm 0.1$}  \\
\midrule
\multirow{3}{*}{\textit{B-III}}
&\multicolumn{1}{c}{ \begin{tabular}[c]{@{}c@{}} LSTM-PE \end{tabular}} &\multicolumn{1}{c}{  73.2 $\pm 0.6$} &\multicolumn{1}{c}{ 85.5 $\pm 0.1$}  \\
&\multicolumn{1}{c}{ \begin{tabular}[c]{@{}c@{}} GRU-PE \end{tabular}} &\multicolumn{1}{c}{  75.7 $\pm 0.7$} &\multicolumn{1}{c}{ 85.7 $\pm 0.1$}   \\
&\multicolumn{1}{c}{ \begin{tabular}[c]{@{}c@{}} GRU-PE (reg.) \end{tabular}} &\multicolumn{1}{c}{  75.7 $\pm 0.5$} &\multicolumn{1}{c}{ 86.1 $\pm 0.1$}   \\
\midrule
\multirow{3}{*}{
\scalebox{0.85}[1.0]{\textit{NODE}}
}
&\multicolumn{1}{c}{ \begin{tabular}[c]{@{}c@{}} $\quad$ODE-LSTM$\quad$ \end{tabular}} &\multicolumn{1}{c}{ 72.9 $\pm 1.0$} &\multicolumn{1}{c}{ 85.4 $\pm 0.1$} \\
&\multicolumn{1}{c}{ \begin{tabular}[c]{@{}c@{}} ODE-GRU \end{tabular}}  &\multicolumn{1}{c}{ 74.9 $\pm 1.2$} &\multicolumn{1}{c}{ 85.5 $\pm 0.1$ } \\
&\multicolumn{1}{c}{ \begin{tabular}[c]{@{}c@{}} ODE-GRU (reg.) \end{tabular}}  &\multicolumn{1}{c}{\textbf{76.5} $\pm 0.2$} &\multicolumn{1}{c}{ \textbf{86.3} $\pm 0.1$} 
\\ \bottomrule
\end{tabular}
\end{center}
\caption{Performance comparison on ZueriCrop data. reg. stands for regularization, see Section \ref{sec:reg} and PE is positional encoding, see Section \ref{sec:baseline}. 
The best score for each metric is printed \textbf{bold}.} 
\label{table:result_swissdata}
\end{table}

\begin{table}[t]
\begin{center}
\setlength{\tabcolsep}{3.6pt}
\begin{tabular}{cllll}
&\multicolumn{1}{c}{\bf Method}
&\multicolumn{1}{c}{\bf \#Parameters}  &\multicolumn{1}{c}{\bf Runtime(ms)} &\multicolumn{1}{c}{\bf Memory(GB)}  
\\
\toprule
\multirow{2}{*}{\textit{B-II}}
&\multicolumn{1}{c}{ \begin{tabular}[c]{@{}c@{}} LSTM-$\delta t$ \end{tabular}} 
&\multicolumn{1}{c}{346k}     &\multicolumn{1}{c}{154.3} &\multicolumn{1}{c}{1.22}\\
&\multicolumn{1}{c}{ \begin{tabular}[c]{@{}c@{}} GRU-$\delta t$ \end{tabular}} 
&\multicolumn{1}{c}{314k}    &\multicolumn{1}{c}{147.7} &\multicolumn{1}{c}{1.16}\\
\midrule
\multirow{2}{*}{
\scalebox{0.85}[1.0]{\textit{NODE}}
}
&\multicolumn{1}{c}{ \begin{tabular}[c]{@{}c@{}} $\quad$ODE-LSTM$\quad$ \end{tabular}}
&\multicolumn{1}{c}{238k}  &\multicolumn{1}{c}{147.4} &\multicolumn{1}{c}{0.97}\\
&\multicolumn{1}{c}{ \begin{tabular}[c]{@{}c@{}} ODE-GRU \end{tabular}} 
&\multicolumn{1}{c}{255k}  &\multicolumn{1}{c}{135.2} &\multicolumn{1}{c}{0.80}
\\ \bottomrule 

\end{tabular}
\end{center}
\caption{Number of trainable parameters, CPU runtime, and GPU memory usage per model (differences between \textit{B-I}, \textit{B-II} and \textit{B-III} are insignificant.) for TUM dataset.
The runtime showed corresponds to the time required to run a forward pass of a batch covering an area of 1 $km^2$ of ground area.
To achieve the best performance, the NODE model needs significantly fewer parameters and memory requirements. The runtime of LSTM is less than of GRU due to the implementation difference of gating functions of two cells, see Appendix.}  
\label{table:params}
\end{table}

\subsection{Dataset Size}
%
%
%
%
The NODE model can be seen as a sort of regularizer that accounts for the continuity of the underlying process and for an isotropic time scale.
In order to show that this regularization, which can be also thought of as a prior, on the continuity of the hidden state (i.e latent dynamics) indeed makes sense and imposes a meaningful inductive bias, we amplify its influence by reducing the amount of training data. We note that this small-data regime is a relevant situation in agricultural remote sensing, where access to ground truth labels is often the bottleneck.

We randomly sub-sample the training set to $10\%$, respectively $1\%$ of its original size, train the models on the reduced set, and evaluate them on the same (full) test set as before.
As expected, performance decreases for both ODE-GRU and the GRU-$\delta t$ baseline if they are presented fewer training samples.
However, the relative advantage due to NODE increases as the model sees less data and must rely more on the prior, as shown in Table~\ref{table:result_low_data}.
This supports our claim that the a-priori assumptions inherent in NODE better match the temporal evolution of agricultural crops, respectively their spectral responses.

%

\begin{table*}[t]
\begin{center}
\tabcolsep=0.106cm
\begin{tabular}{lllllllll}
\multicolumn{1}{c}{} &&\multicolumn{3}{c}{\bf F1-score (\%)} & & \multicolumn{3}{c}{\bf Accuracy (\%)} \\
%
\multicolumn{1}{c}{\bf Method}  &&\multicolumn{1}{c}{100\%}  &\multicolumn{1}{c}{10\%} &\multicolumn{1}{c}{ 1\%}  &&\multicolumn{1}{c}{100\%}  &\multicolumn{1}{c}{ 10\%} &\multicolumn{1}{c}{ 1\%} 
\\ \toprule
\multicolumn{1}{c}{ \begin{tabular}[c]{@{}c@{}} GRU-$\delta  t$ \end{tabular} }    &&78.2 $\pm 0.7$ & 70.1 $\pm 0.5$ & 49.4 $\pm 0.8$ && 87.6 $\pm 0.1$ & 84.9 $\pm 0.1$ & 79.3 $\pm 0.4$
\\
\multicolumn{1}{c}{\begin{tabular}[c]{@{}c@{}} ODE-GRU \end{tabular}}  &&\textbf{79.9} $\pm 0.2$ &\textbf{72.2} $\pm 0.6$ &\textbf{52.5} $\pm 2.3$ &&\textbf{88.1} $\pm 0.1$ &\textbf{85.5} $\pm 0.1$ & \textbf{80.3} $\pm 0.2$
\\ \cmidrule(lr){1-1}\cmidrule(lr){3-5} \cmidrule(lr){7-9}
\multicolumn{1}{c}{ \begin{tabular}[c]{@{}c@{}} \it Difference \end{tabular}}   &&+1.7 &+2.1 &+3.1 &&+0.5 &+0.6 &+1.0   
\\ \bottomrule
\end{tabular}
\end{center}
\caption{Performance comparison for small-data regime (TUM dataset), using 100\%, 10\%, and 1\% of the available training sequences. 
}
\label{table:result_low_data}
\end{table*}

\subsection{Early Classification}
\begin{figure}[t]
    \centering
    \includegraphics[width=0.8\columnwidth]{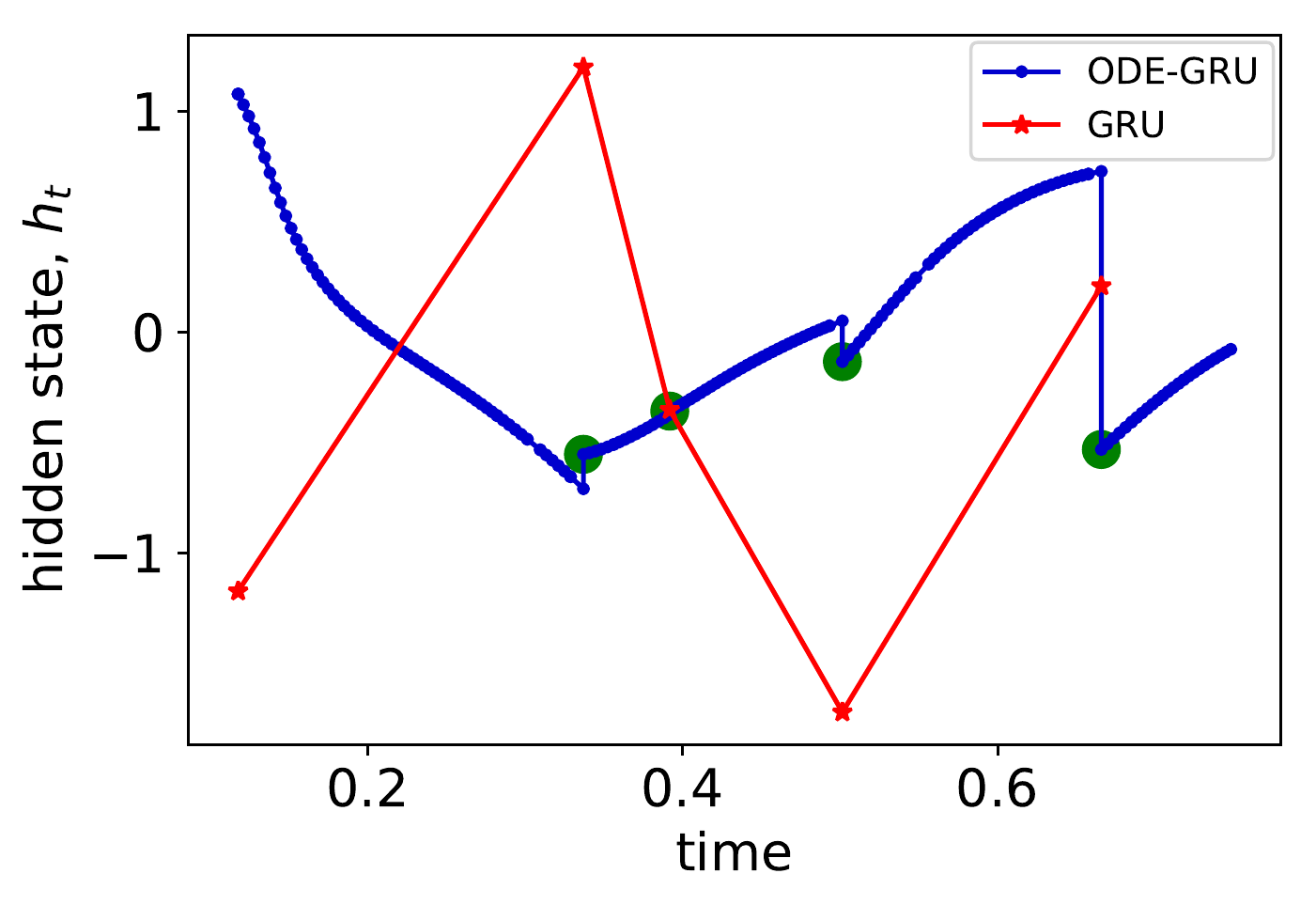}
    \caption{Hidden state trajectories (reduced to 1 dimension with Principal Component Analysis). The blue and green markers denote \emph{prediction} and \emph{update} steps of ODE-GRU, respectively. Red markers denote GRU updates. Note that small step size in \emph{prediction} is for visualization purposes, larger step size is used in the experiments.}
    \label{fig:trajectory}
\end{figure}

The architecture as proposed in Section~\ref{Method} is capable of extrapolating the hidden trajectory beyond the last data point in time, whereas conventional RNNs (including GRU) cannot extrapolate without observations -- this is illustrated on the right end of the trajectories in Fig.~~\ref{fig:trajectory}.
In the next experiment we investigate this difference, by comparing our ODE-GRU to the GRU-$\delta t$ baseline on the task of early classification.
That task corresponds to the practically important scenario of forecasting the area of each crop from a shorter time series covering only the early part of the growing season, before most crops have been harvested.
In the context of agriculture, forecasting is a practically important scenario, for instance to ensure food security or to inform policies for sustainable agriculture~\cite{marc_early_classification}.
To simulate that setting, we truncate the time series of the test set and keep only the leading $75\%$, respectively $50\%$ of all time steps.
For the conventional GRU, this means that we have to classify from only the early part of the growing season, whereas with ODE-GRU we can simply let the prediction step run beyond the last time step without observations. We also tried to extrapolate with the conventional GRU, by feeding it the global channel-wise mean values as input, but this did not improve the scores.
Obviously classification performance drops if only a part of the seasonal cycle is shown to the model (Table~\ref{table:early}). However, the relative improvement due to NODE increases quite significantly as the time series ends earlier.
This demonstrates the value of explicitly modelling the latent trajectories and being able to extrapolate them into the future beyond the last observation.

\begin{table*}[t]
\begin{center}
\tabcolsep=0.106cm
\begin{tabular}{lllllllll}
\multicolumn{1}{c}{} &&\multicolumn{3}{c}{\bf F1-score (\%)} && \multicolumn{3}{c}{\bf Accuracy (\%)}  \\
\multicolumn{1}{c}{\bf Method}  &&\multicolumn{1}{c}{100\%}  &\multicolumn{1}{c}{75\%} &\multicolumn{1}{c}{ 50\%}   &&\multicolumn{1}{c}{100\%}  &\multicolumn{1}{c}{ 75\%} &\multicolumn{1}{c}{ 50\%} 
\\ \toprule
\multicolumn{1}{c}{ \begin{tabular}[c]{@{}c@{}} GRU-$\delta  t$ \end{tabular} }  &&78.2 ${\pm0.7}$ &51.2 ${\pm1.8}$ &19.2 ${\pm1.3}$ &&87.6 ${\pm0.1}$  & 74.9 ${\pm0.0}$ &46.9 ${\pm1.5}$   
\\
\multicolumn{1}{c}{\begin{tabular}[c]{@{}c@{}} ODE-GRU \end{tabular}}   &&\textbf{79.9} ${\pm0.2}$ &\textbf{54.5} ${\pm0.6}$ &\textbf{24.3} ${\pm0.5}$  &&\textbf{88.1} ${\pm0.1}$ &\textbf{79.8} ${\pm0.2}$ &\textbf{60.4} ${\pm0.7}$
\\ \cmidrule(lr){1-1} \cmidrule(lr){3-5} \cmidrule(lr){7-9}
\multicolumn{1}{c}{ \begin{tabular}[c]{@{}c@{}} \it Difference \end{tabular}}   &&+1.7 &+3.3 &+5.1  && +0.5 &+4.9 &+13.5 
\\ \bottomrule
\end{tabular}
\end{center}
\caption{Performance comparison for early classification (TUM dataset). 
}
\label{table:early}
\end{table*}

\subsection{Sparser Time Series}
In this experiment, the input image time series are also shortened, but this time by randomly
subsampling along the time dimension rather than truncation. I.e., the model receives fewer observations ($100/75/50/25$\% of the original number) per time series, but those are still distributed over the entire phenological cycle, just sparser.
Please note that, in this experiment, we do not subsample the dataset i.e., the entire training and test datasets are still presented to the models. Such approach can be considered as an attempt to artificially add more clouds to the time series in the same dataset.
%
%
%
%
%
%
When training and test data have progressively higher (but matched) sparsity, performance decreases monotonically for both models as expected, see Table~\ref{table:missingness}. 
%
ODE-GRU consistently outperforms the baseline, but also performance gap increases as sparsity increases, showing that the stronger dynamical modelling power of NODE manages to preserve a meaningful hidden state across longer temporal gaps.
%
%
%
%
The better handling of sparse time series with few input images is consistent with our hypothesis that NODE can represent the temporal evolution more faithfully and is therefore able to bridge longer and more irregular data gaps. \\

%

\begin{table*}[t]
\begin{center}
\tabcolsep=0.106cm
\begin{tabular}{lllllllllll}
\multicolumn{1}{c}{} &&\multicolumn{4}{c}{\bf F1-score (\%)} && \multicolumn{4}{c}{\bf Accuracy (\%)}  \\
\multicolumn{1}{c}{\bf Method}  &&\multicolumn{1}{c}{100\%}  &\multicolumn{1}{c}{75\%} &\multicolumn{1}{c}{ 50\%} &\multicolumn{1}{c}{ 25\%}  &&\multicolumn{1}{c}{100\%}  &\multicolumn{1}{c}{ 75\%} &\multicolumn{1}{c}{ 50\%} &\multicolumn{1}{c}{ 25\%} 
\\ \toprule
\multicolumn{1}{c}{ \begin{tabular}[c]{@{}c@{}} GRU-$\delta  t$ \end{tabular} }  &&78.2 ${\pm0.7}$ &75.9 ${\pm0.4}$ &70.5 ${\pm0.2}$ &52.6 ${\pm0.5}$    &&87.6 ${\pm0.1}$  & 86.1 ${\pm0.1}$ &85.0 ${\pm0.1}$  &77.8 ${\pm0.1}$   
\\
\multicolumn{1}{c}{\begin{tabular}[c]{@{}c@{}} ODE-GRU \end{tabular}}   &&\textbf{79.9} ${\pm0.2}$ &\textbf{77.6} ${\pm0.4}$ &\textbf{73.5} ${\pm0.2}$ &\textbf{58.5} ${\pm0.1}$  &&\textbf{88.1} ${\pm0.1}$ &\textbf{87.6} ${\pm0.0}$ &\textbf{86.2} ${\pm0.1}$ &\textbf{80.8} ${\pm0.0}$   
\\ \cmidrule(lr){1-1} \cmidrule(lr){3-6} \cmidrule(lr){7-11}
\multicolumn{1}{c}{ \begin{tabular}[c]{@{}c@{}} \it Difference \end{tabular}}   &&+1.7 &+1.7 &+3.0  &+5.9  && +0.5 &+1.5 &+1.2 &+3.0 
\\ \bottomrule
\end{tabular}
\end{center}
\caption{Performance comparison for sparser time series (TUM dataset) with input observations subsampled to 100\%, 75\%, 50\%, 25\% of the original density. 
}
\label{table:missingness}
\end{table*}

\subsection{Sensitivity Analysis for Regularization}
\begin{figure}[t]
    \centering
    \includegraphics[width=0.9\columnwidth]{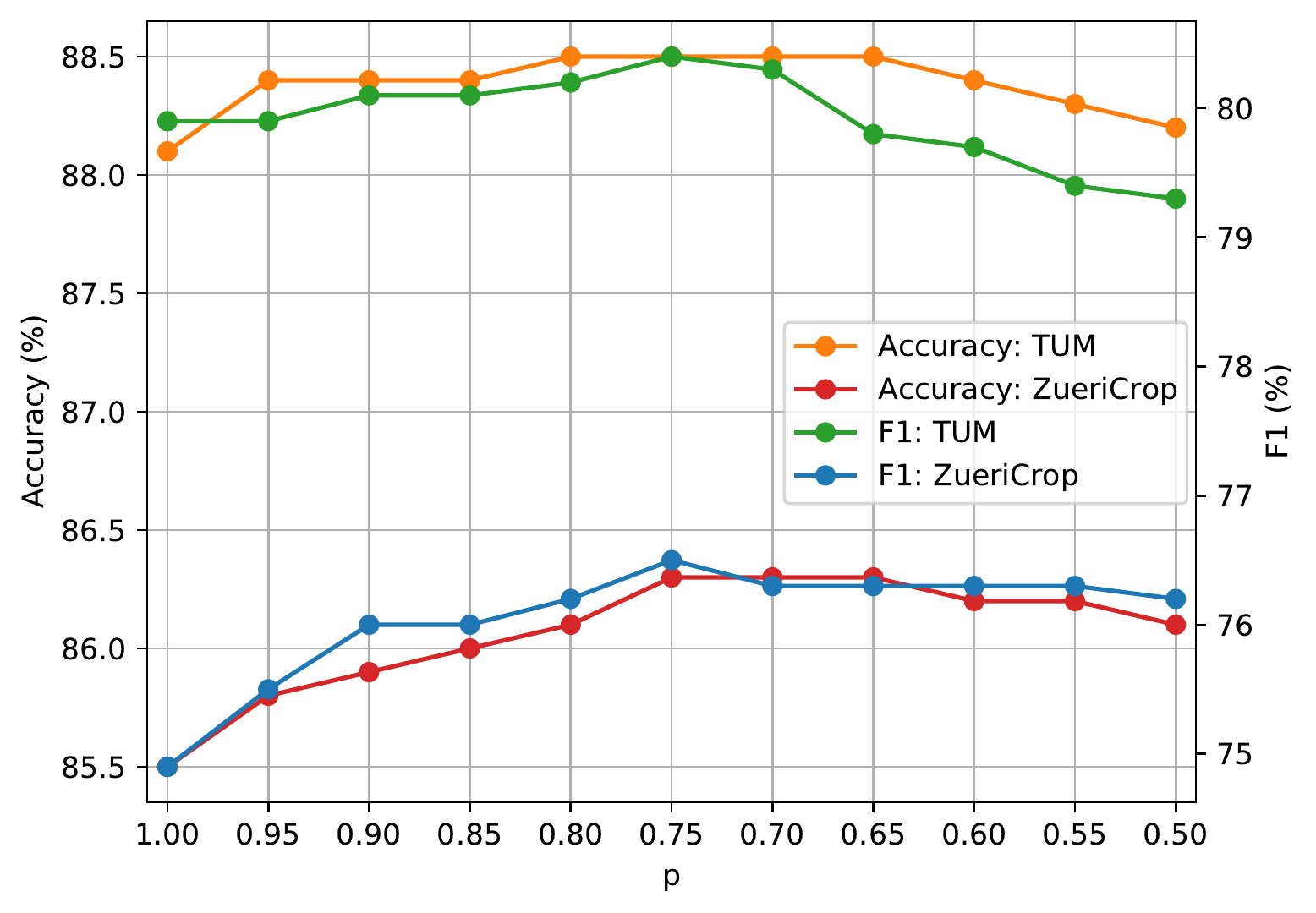}
    \caption{Performance of ODE-GRU on TUM and ZueriCrop datasets, as a function of the regularization hyper-parameter $p$. C.f.\ Section~\ref{sec:reg}.}
    \label{fig:sensitivity_p}
\end{figure}

In Section~\ref{sec:reg} we have described a regularization scheme where training sequences are randomly subsampled with a factor $p\!\in\![0\hdots 1]$ ($p=1$ means no regularization). Figure \ref{fig:sensitivity_p} displays the performance of the proposed ODE-GRU model as a function of the hyper-parameter $p$, evaluated on both datasets. Randomly omitting a reasonable fraction of the input time series improves generalization a bit, for both datasets the best scores are achieved with $p=0.75$, we further observed that  $p$ values smaller than $0.5$ lead to a performance drop. See also entries denoted by "(reg.)" in Tables~\ref{table:result_tum} and~\ref{table:result_swissdata}. Note the relation to dropout regularization, where individual neurons are omitted at random.

%% file: 06_conclusion.tex
We have studied the use of neural ordinary differential equations (NODE) in combination with conventional recurrent units to assimilate image observations, in order to model the latent dynamics of spectral signatures over time.
Our target application is crop classification from optical satellite time series, with missing data due to cloud cover.
We have shown that the ODE-RNN model performs consistently better than conventional recurrent network architectures that lack an explicit, isotropic notion of time.
Under favourable conditions with frequent revisits and low to moderate cloud cover conventional RNN models already perform rather well, still we have observed consistent improvements.
The NODE prior becomes particularly useful under challenging conditions, for instance when only little training data is available, when the available time series are sparse, and when the time series cover only a part of the growth cycle.
In those situations NODE exhibits significant benefits in our experiments, reaching up to 10\% higher F1-score for very sparse time series and up to 20\% higher overall accuracy for the early classification scenario.
%
As a side effect, we also found that the ability to represent irregularly sampled time series and assimilate observations at arbitrary points in time allows for a convenient form of data augmentation by randomly dropping some observations from the training sequences.

So far, we have chosen the most straight-forward way to update the hidden dynamics of NODE when an observation becomes available.
However, this procedure leads to discontinuous jumps of the hidden state with each update. We suspect that these jumps adversely affect the model, because they violate the ODE's preference for a continuously evolving state vector. In future work we plan to address that limitation, perhaps through tighter integration between the prediction and update mechanisms or regularization of the update step.

Beyond our specific application, we believe that the NODE framework has great potential for time series analysis in remote sensing and Earth observation, where data gaps are ubiquitous, not only due to clouds, but also caused by irregular acquisition patterns in space or time, sensor failures and replacements, transmission limits, etc.

%% file: biographies/bios.tex
\begin{IEEEbiography}[{\includegraphics[width=1.0in,height=1.25in,clip,keepaspectratio]{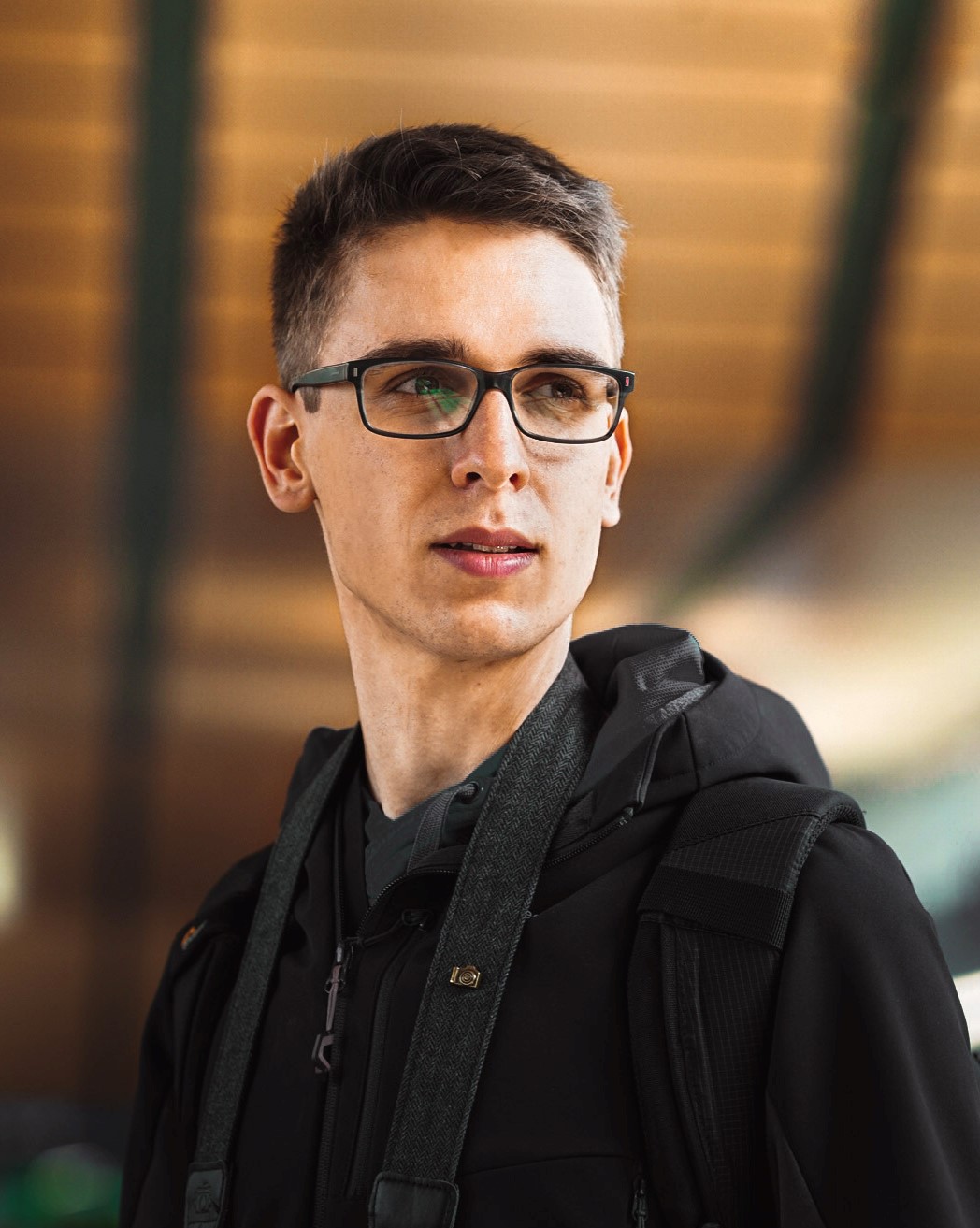}}]{Nando Metzger}
received his BSc degrees in geomatics engineering at ETH Zurich in 2019. In 2021, he graduated from ETH Zurich and received his master's degree in geomatics engineering, focusing on photogrammetry and remote sensing. His academic interests include computer vision, deep learning, and their applications to remote sensing data.
\end{IEEEbiography}

\begin{IEEEbiography}[{\includegraphics[width=1.0in,height=1.25in,clip,keepaspectratio]{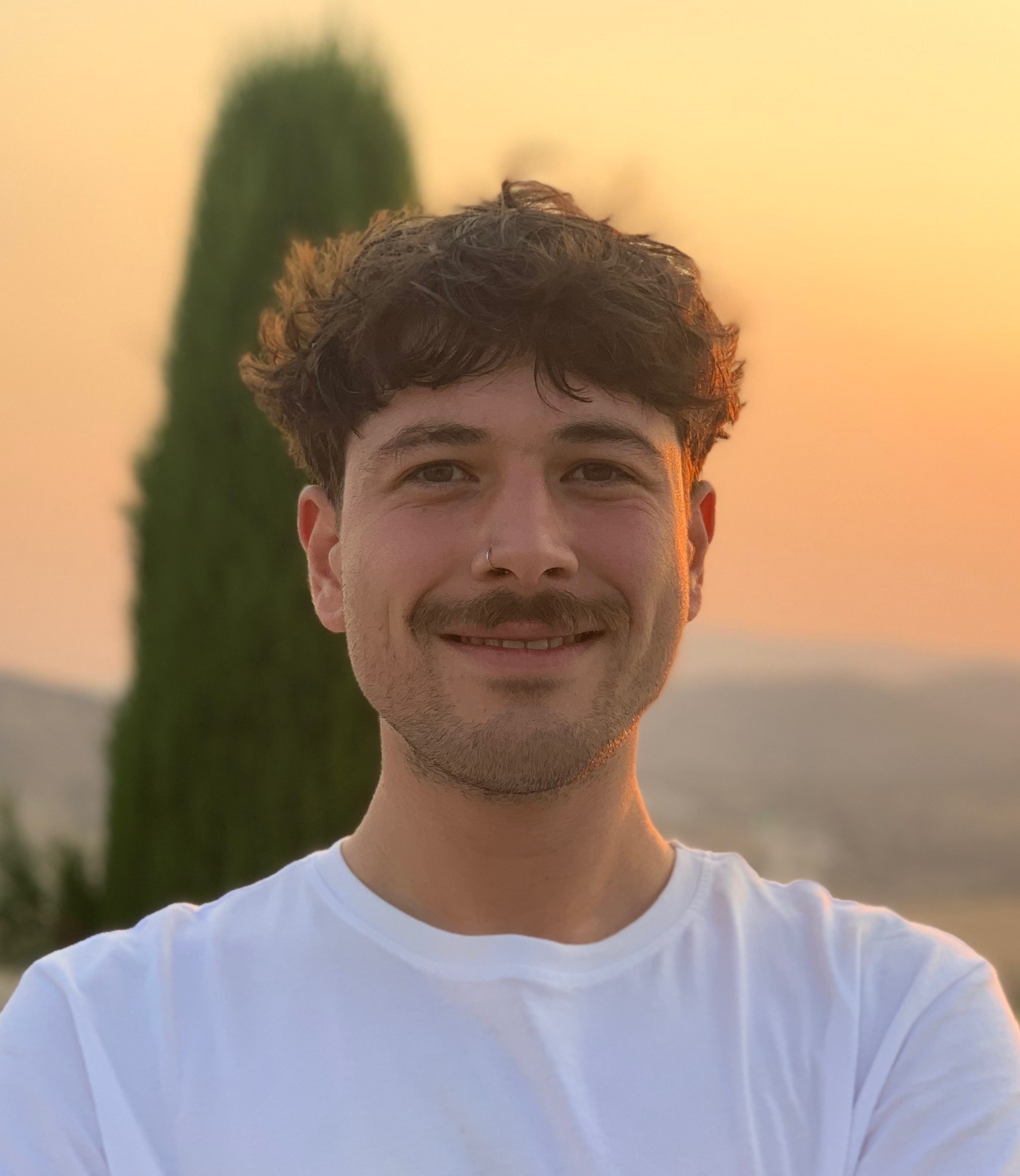}}]{Mehmet Ozgur Turkoglu}
received his BSc degrees in both electrical engineering and physics from Bogazici University in 2016. He studied a master's in electrical engineering with a specialization in computer vision at the University of Twente. He is a PhD candidate in the EcoVision group at ETH Z\"urich since 2018. His research interests include computer vision, deep learning and their applications to remote sensing data. He is particularly interested in deep sequence modeling of time-series data.
\end{IEEEbiography}

\begin{IEEEbiography}[{\includegraphics[width=1.0in,height=1.25in,clip,keepaspectratio]{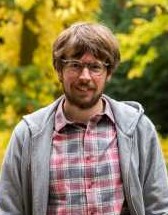}}]{Stefano D'Aronco}
received his BS and MS degrees in electronic engineering from the Università degli studi di Udine, in 2010 and 2013 respectively. He then joined the Signal Processing Laboratory (LTS4) in 2014 as a PhD student under the supervision of Prof. Pascal Frossard. He received his PhD in Electrical Engineering from École Polytechnique Fédérale de Lausanne in 2018. He is Postdoctoral researcher in the EcoVision group at ETH Z\"urich since 2018. His research interests include several machine learning topics, such as Bayesian inference method and deep learning, with particular emphasis on applications related to remote sensing an environmental monitoring.
\end{IEEEbiography}

\begin{IEEEbiography}[{\includegraphics[width=1.0in,height=1.25in,clip,keepaspectratio]{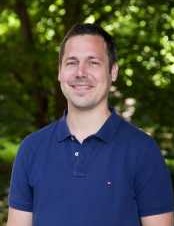}}]{Jan Dirk Wegner}
is associate professor at University of Zurich and head of the EcoVision Lab at ETH Zurich. Jan was PostDoc (2012-2016) and senior scientist (2017-2020) in the Photogrammetry and Remote Sensing group at ETH Zurich after completing his PhD (with distinction) at Leibniz Universität Hannover in 2011. He was granted multiple awards, among others an ETH Postdoctoral fellowship and the science award of the German Geodetic Commission. Jan was selected for the WEF Young Scientist Class 2020 as one of the 25 best researchers world-wide under the age of 40 committed to integrating scientific knowledge into society for the public good. Jan is vice-president of ISPRS Technical Commission II, chair of ISPRS II/WG 6 "Large-scale machine learning for geospatial data analysis", director of the PhD graduate school "Data Science" at University of Zurich, and his professorship is part of the Digital Society Initiative at University of Zurich. 
\end{IEEEbiography}

\begin{IEEEbiography}[{\includegraphics[width=1.0in,height=1.25in,clip,keepaspectratio]{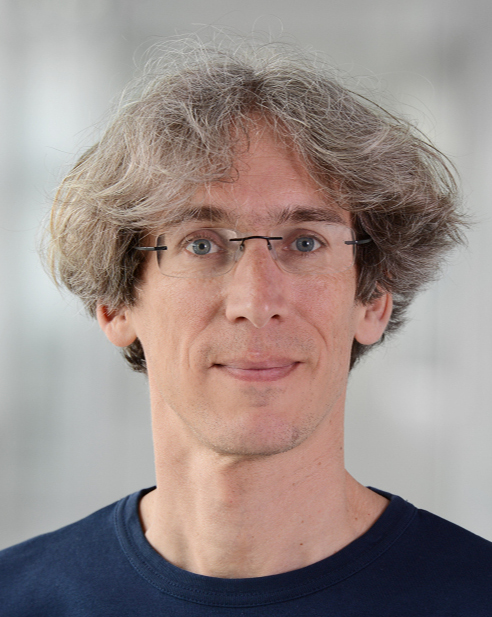}}]{Konrad Schindler}
(M'05–SM'12) received the Diplomingenieur (M.Tech.) degree from Vienna University of Technology, Vienna, Austria, in 1999, and the Ph.D.\ degree from Graz University of Technology, Graz, Austria, in 2003.
He was a Photogrammetric Engineer in the private industry and held researcher positions at Graz University of Technology, Monash University, Melbourne, VIC, Australia, and ETH Z\"urich, Z\"urich, Switzerland. He was an Assistant Professor of Image Understanding with TU Darmstadt, Darmstadt, Germany, in 2009. Since 2010, he has been a Tenured Professor of Photogrammetry and Remote Sensing with ETH Z\"urich. His research interests include computer vision, photogrammetry, and remote sensing.
\end{IEEEbiography}
\vfill

%% file: 08_appendix.tex
\onecolumn
\section{Effect of spatial context aggregation}
We conducted an additional experiment with ODE-GRU model on ZueriCrop dataset using the center pixel (1x1 patch) instead of using 9 pixels (3x3 patch), see Table~\ref{table:center} for the performance comparison. Using a 3x3 patch improves the model performance especially in terms of F1 score but it slightly increases the memory cost, too. We find that the additional number of parameters and runtime are insignificant.
\begin{table*}[ht]
\begin{center}
\begin{tabular}{cllllll}
&\multicolumn{1}{c}{\bf Patch Size}
&\multicolumn{1}{c}{\bf \#Parameters}  
&\multicolumn{1}{c}{\bf Runtime (ms)} 
&\multicolumn{1}{c}{\bf Memory (GB)} 
&\multicolumn{1}{c}{\bf F1 (\%)} 
&\multicolumn{1}{c}{{\bf Accuracy (\%)}}
\\
\toprule
&\multicolumn{1}{c}{ \begin{tabular}[c]{@{}c@{}} 1x1 \end{tabular}} 
&\multicolumn{1}{c}{241k}  
&\multicolumn{1}{c}{264.6} 
&\multicolumn{1}{c}{1.07}
&\multicolumn{1}{c}{73.1 $\pm 0.3$}
&\multicolumn{1}{c}{85.2 $\pm 0.1$}\\
&\multicolumn{1}{c}{ \begin{tabular}[c]{@{}c@{}} 3x3 \end{tabular}} 
&\multicolumn{1}{c}{248k}    
&\multicolumn{1}{c}{267.8} 
&\multicolumn{1}{c}{1.14}
&\multicolumn{1}{c}{\textbf{74.9 $\pm 1.2$}}
&\multicolumn{1}{c}{\textbf{85.5 $\pm 0.1$}}
\\ \bottomrule 

\end{tabular}
\end{center}
\caption{Effect of input patch size for ODE-GRU model on ZueriCrop dataset. Using 3x3 patch (9 pixels) instead of 1x1 patch (center pixel) improves the model performance while slightly increasing memory usage. We find that the additional number of parameters and runtime are insignificant.} 
\label{table:center}
\end{table*}

\section{Hyper-parameters Settings}
We provide detailed hyper-parameters setting of the models in Table~\ref{table:params_tum} and Table~\ref{table:params_swiss} for TUM and ZueriCrop datasets, respectively. Code link for the other baselines are give in Table~\ref{table:source}.

\begin{table*}[ht]
\begin{center}
\begin{tabular}{cllllllll}
&\multicolumn{1}{c}{\bf Method}
&\multicolumn{1}{c}{{\bf Hidden Dim}}
&\multicolumn{1}{c}{{\bf Layers in $f_{\theta}$}}
&\multicolumn{1}{c}{{\bf Units $f_{\theta}$}}
&\multicolumn{1}{c}{{\bf Gating Layers}}
&\multicolumn{1}{c}{{\bf Gating Units}}
&\multicolumn{1}{c}{\bf \#Parameters}  
\\
\toprule
%
%
%
%
%
\multirow{2}{*}{\textit{B-II}}
&\multicolumn{1}{c}{ \begin{tabular}[c]{@{}c@{}} LSTM-$\delta t$ \end{tabular}} 
&\multicolumn{1}{c}{{150}}
&\multicolumn{1}{c}{{-}}
&\multicolumn{1}{c}{-} 
&\multicolumn{1}{c}{{1}}
&\multicolumn{1}{c}{{-}} 
&\multicolumn{1}{c}{{346k}}\\
&\multicolumn{1}{c}{ \begin{tabular}[c]{@{}c@{}} GRU-$\delta t$ \end{tabular}} 
&\multicolumn{1}{c}{{150}}
&\multicolumn{1}{c}{{-}}
&\multicolumn{1}{c}{-}    
&\multicolumn{1}{c}{{2}} 
&\multicolumn{1}{c}{{100}} 
&\multicolumn{1}{c}{{314k}}\\
\midrule
\multirow{2}{*}{
\scalebox{0.85}[1.0]{\textit{NODE}}
}
&\multicolumn{1}{c}{ \begin{tabular}[c]{@{}c@{}} $\quad$ODE-LSTM$\quad$ \end{tabular}}
&\multicolumn{1}{c}{{85}} 
&\multicolumn{1}{c}{{2}}
&\multicolumn{1}{c}{255}  
&\multicolumn{1}{c}{{1}}
&\multicolumn{1}{c}{{-}} 
&\multicolumn{1}{c}{{238k}}\\
&\multicolumn{1}{c}{ \begin{tabular}[c]{@{}c@{}} ODE-GRU \end{tabular}} 
&\multicolumn{1}{c}{{80}}
&\multicolumn{1}{c}{{2}}
&\multicolumn{1}{c}{255} 
&\multicolumn{1}{c}{{2}}
&\multicolumn{1}{c}{{100}}
&\multicolumn{1}{c}{{255k}}
\\ \bottomrule 

\end{tabular}
\end{center}
\caption{TUM dataset: Parameter settings for baseline, and the proposed NODE models.}
\label{table:params_tum}
\end{table*}

\begin{table*}[ht]
\begin{center}
\begin{tabular}{cllllllll}
&\multicolumn{1}{c}{\bf Method}
&\multicolumn{1}{c}{{\bf Hidden Dim}}
&\multicolumn{1}{c}{{\bf Layers in $f_{\theta}$}}
&\multicolumn{1}{c}{{\bf Units $f_{\theta}$}}
&\multicolumn{1}{c}{{\bf Gating Layers}}
&\multicolumn{1}{c}{{\bf Gating Units}}
&\multicolumn{1}{c}{\bf \#Parameters}  
\\
\toprule
%
%
%
%
%
\multirow{2}{*}{\textit{B-II}}
&\multicolumn{1}{c}{ \begin{tabular}[c]{@{}c@{}} LSTM-$\delta t$ \end{tabular}} 
&\multicolumn{1}{c}{{130}}
&\multicolumn{1}{c}{{-}}
&\multicolumn{1}{c}{-}
&\multicolumn{1}{c}{{1}}
&\multicolumn{1}{c}{{-}}
&\multicolumn{1}{c}{{293k}}\\
&\multicolumn{1}{c}{ \begin{tabular}[c]{@{}c@{}} GRU-$\delta t$ \end{tabular}} 
&\multicolumn{1}{c}{{120}}
&\multicolumn{1}{c}{{-}}
&\multicolumn{1}{c}{-}  
&\multicolumn{1}{c}{{2}}
&\multicolumn{1}{c}{{80}}
&\multicolumn{1}{c}{{272k}}\\
\midrule
\multirow{2}{*}{
\scalebox{0.85}[1.0]{\textit{NODE}}
}
&\multicolumn{1}{c}{ \begin{tabular}[c]{@{}c@{}} $\quad$ODE-LSTM$\quad$ \end{tabular}}
&\multicolumn{1}{c}{{85}}
&\multicolumn{1}{c}{{2}}
&\multicolumn{1}{c}{255} 
&\multicolumn{1}{c}{{1}}
&\multicolumn{1}{c}{{-}}
&\multicolumn{1}{c}{{251k}}\\
&\multicolumn{1}{c}{ \begin{tabular}[c]{@{}c@{}} ODE-GRU \end{tabular}} 
&\multicolumn{1}{c}{{80}} 
&\multicolumn{1}{c}{{2}}
&\multicolumn{1}{c}{230}
&\multicolumn{1}{c}{{2}}
&\multicolumn{1}{c}{{80}} 
&\multicolumn{1}{c}{{248k}}
\\ \bottomrule 

\end{tabular}
\end{center}
\caption{ZueriCrop dataset: Parameter settings for baseline, and the proposed NODE models.}
\label{table:params_swiss}
\end{table*}

\begin{table*}[ht]
\begin{center}
\tabcolsep=0.05cm
\begin{tabular}{ |c|c|} 
 \hline
 \textbf{Method} &  \textbf{Link} \\ 
 \hline
 TCN &  \url{https://github.com/charlotte-pel/temporalCNN}\\ 
 \hline
 Transformer &  \url{https://github.com/VSainteuf/lightweight-temporal-attention-pytorch}\\
 \hline
\end{tabular}
\end{center}
\caption{Code sources of baseline methods. }
\label{table:source}
\end{table*}